\documentclass[11pt,a4paper, english]{article}
\pdfoutput=1
\usepackage{graphicx}
\usepackage{subcaption}
\usepackage{longtable}
\usepackage{enumitem}
\usepackage{xspace}
\usepackage{soul} % for strikeout
\usepackage{array} % for column width control
\usepackage[margin=1in]{geometry} % Adjust the margin here
\usepackage{booktabs}
\usepackage{url}
\usepackage{multirow}
\usepackage{pgfplots}
\usepackage{hyperref}
\pgfplotsset{compat=1.14}

\newcommand{\team}{Magentic-One\xspace}
\newcommand{\agbench}{AutoGenBench}

\title{\team: A Generalist  Multi-Agent System\\ for Solving Complex Tasks}
\author{  
$\star$ Adam Fourney, Gagan Bansal, Hussein Mozannar, Cheng Tan $\star$\\ $\dagger$  Eduardo Salinas, Erkang (Eric) Zhu, Friederike Niedtner,  Grace Proebsting, \\   Griffin Bassman, Jack Gerrits,  Jacob Alber,  
 Peter Chang, \\ Ricky Loynd, Robert West, Victor Dibia $\dagger$\\   $\diamond$  Ahmed Awadallah, Ece Kamar,   Rafah Hosn,   Saleema Amershi $\diamond$\\ \\  {\bf Microsoft Research AI Frontiers} }

\date{}

\begin{document}

\maketitle
\begingroup
\renewcommand\thefootnote{}
\footnotetext{$\star$: Research Leads, $\dagger$: Core Contributors, $\diamond$: Program Leads. Contact: \href{mailto:magentic-one@microsoft.com}{magentic-one@microsoft.com}}
\addtocounter{footnote}{0}
\endgroup

\vspace{-0.5cm}

\begin{figure}[!h]
    \centering
    \vspace{-.5em}
    \includegraphics[width=.95\linewidth]{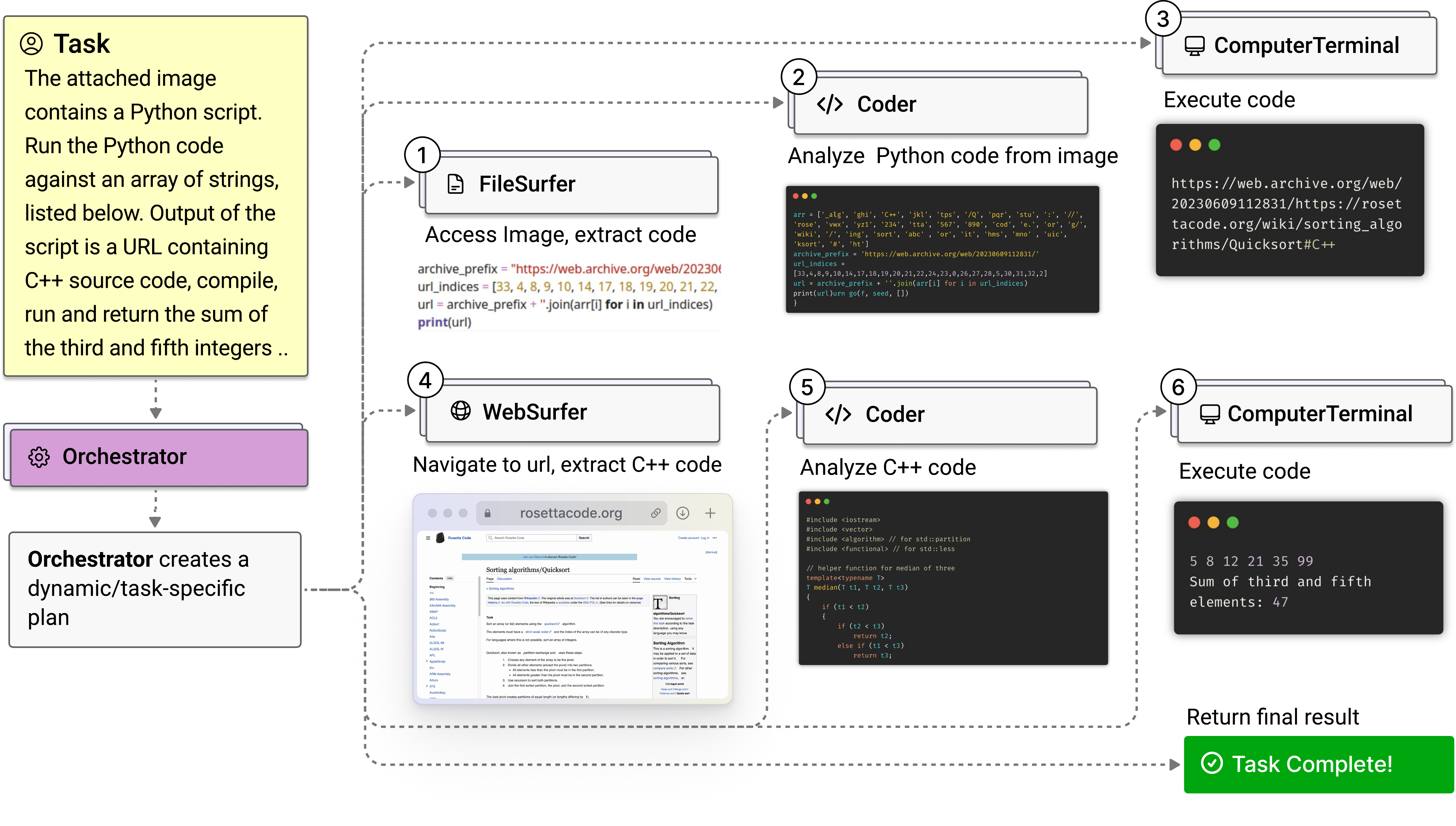}
    \vspace{-.5em}
    \caption{An illustration of the \team mutli-agent team completing a complex task from the GAIA benchmark. \team's Orchestrator agent creates a plan, delegates tasks to other agents, and tracks progress towards the goal, dynamically revising the plan as needed. The Orchestrator can delegate tasks to a FileSurfer agent to read and handle files, a WebSurfer agent to operate a web browser, or a Coder or Computer Terminal agent to write or execute code, respectively.}
    \label{fig:magentic-example}
\end{figure} 
\begin{abstract}

\noindent
Modern AI agents, driven by advances in large foundation models, promise to enhance our productivity and transform our lives by augmenting our knowledge and capabilities. To achieve this vision, AI agents must effectively plan, perform multi-step reasoning and actions, respond to novel observations, and recover from errors, to successfully complete complex tasks across a wide range of scenarios. In this work, we introduce {\em \team}, a high-performing open-source agentic system for solving such tasks. \team uses a multi-agent architecture where a lead agent, the \emph{Orchestrator}, plans, tracks progress, and re-plans to recover from errors. Throughout task execution, the Orchestrator also directs other specialized agents to perform tasks as needed, such as operating a web browser, navigating local files, or writing and executing Python code. Our experiments show that \team achieves statistically competitive performance to the state-of-the-art on three diverse and challenging agentic benchmarks: GAIA, AssistantBench, and WebArena. Notably, \team achieves these results without modification to core agent capabilities or to how they collaborate, demonstrating progress towards the vision of \emph{generalist agentic systems}. Moreover, \team's modular design allows agents to be added or removed from the team without additional prompt tuning or training, easing development and making it extensible to future scenarios. We provide an open-source implementation of \team, and we include \agbench{}, a standalone tool for agentic evaluation. \agbench{} provides built-in controls for repetition and isolation to run agentic benchmarks in a rigorous and contained manner -- which is important when agents' actions have side-effects. \team, \agbench{} and detailed empirical performance evaluations of \team,  including ablations and error analysis are available at \url{https://aka.ms/magentic-one}.
\end{abstract}

\section{Introduction}

Recent advances in artificial intelligence and foundation models are driving a renewed interest in \emph{agentic systems} that can perceive, reason, and act in the world to complete tasks on our behalf \cite{russellnorvig1996, Wooldridge1995}. These systems promise to enhance our productivity by relieving us from mundane and laborious tasks, and revolutionize our lives by augmenting our knowledge and capabilities \cite{Jennings1998Applications,Wang2023survey, cheng2024exploring}. 
By leveraging the powerful reasoning and generative capabilities of large language models (LLMs), agentic systems are already making strides in fields like software engineering  \cite{yang2024swe,wang2024opendevinopenplatformai}, data analysis \cite{cao2024spider2vfarmultimodalagents}, scientific research \cite{lu2024ai,d2024marg} and web navigation \cite{zhou2024webarenarealisticwebenvironment,zhang2024webpilotversatileautonomousmultiagent}.

Realizing the vision of agentic systems to transform our lives requires these systems to not only achieve high performance in specific domains, but also to generalize to the diverse range of tasks people may encounter throughout their day-to-day work and personal lives. In this paper, we take steps towards creating such a \emph{generalist agentic system} by introducing \emph{\team}.\footnote{The name \team{} is a combination of the words \textit{\textbf{m}ulti} and \textit{\textbf{agentic}}.} \team uses a team ofagents, each specializing in generally-useful skills, such as: operating a web browser, handling files, and executing code. The team is directed by an Orchestrator agent which guides progress towards a high-level goal by iteratively planning, maintaining working memory of progress, assigning tasks to other agents, and retrying upon encountering errors. The Orchestrator uses two \emph{structured ledgers} to achieve this and also to decide which agent should take the next action. 
Together, \team{}'s agents achieve strong performance on multiple challenging agentic benchmarks. 
Figure \ref{fig:magentic-example} shows an example of \team solving one such benchmark task that requires multiple steps and diverse tools. 

Key to \team's performance is its modular and flexible multi-agent approach \cite{MASaSurvey2000,Messing2002AnIT,Tambe1998AgentTeams,guo2024large,talebirad2023multiagentcollaborationharnessingpower},  implemented via the AutoGen\footnote{\url{https://github.com/microsoft/autogen}} framework \cite{wu2023autogen}. The multi-agent paradigm offers numerous advantages over monolithic single-agent approaches \cite{MASaSurvey2000,Tambe1998AgentTeams,cheng2024exploring,xi2023risepotentiallargelanguage}, which we believe makes it poised to become the leading paradigm in agentic development. For example, encapsulating distinct skills in separate agents simplifies development and facilitates reusability, akin to object-oriented programming. \team's specific design further supports easy adaptation and extensibility by enabling agents to be added or removed without altering other agents, or the overall workflow, unlike single-agent systems that often struggle with constrained and inflexible workflows.

To rigorously evaluate \team's performance, we introduce \emph{\agbench{}}, an extensible standalone tool for running agentic benchmarks. \agbench{}'s design enables repetition, isolation, and strong controls over initial conditions, so as to accommodate the variance of stochastic LLM calls, and to isolate the side-effects of agents taking actions.  Using \agbench{}, we evaluated \team on three agentic benchmarks. We observed task-completion rates of 38\% on GAIA \cite{mialon2023gaiabenchmarkgeneralai} and 32.8\% on WebArena \cite{zhou2024webarenarealisticwebenvironment}; and attained an accuracy of 27.7\% on AssistantBench \cite{yoran2024assistantbenchwebagentssolve}. These results place  \team in a strong position, where it is statistically competitive with other state-of-the-art (SOTA) systems, including those that are specialized for a given benchmark.  
Follow-up ablation experiments and in-depth error analyses reveal the additive value of each agent to \team's performance, and highlight opportunities for further improvement. 

In summary, we contribute: 
\begin{enumerate}
    \item \emph{\team}, a generalist multi-agent team with an open-source implementation. The team consists of five agents: a Coder, Computer Terminal, File Surfer, Web Surfer, and Orchestrator. Different agents can operate relevant tools such as stateful Web and file browsers, as well as command line and Python code executors. The Orchestrator  performs several functions to guide progress towards accomplishing a high-level goal: it formulates a plan, maintains structured working memory of progress, directs tasks to other agents, restarts and resets upon stalling, and determines task completion. 
    \item \emph{\agbench{}}, a standalone tool for evaluating systems on agentic benchmarks, also made available open-source.\footnote{\url{https://aka.ms/agbench}} \agbench{} handles  configuring, running, and reporting performance of agentic solutions while ensuring that all experiments start with well-known initial conditions, and that agents cannot interfere with one another across runs. 
    \item Experimental results and analyses of \team's performance on the GAIA, WebArena, and AssistantBench benchmarks, demonstrating strong task completion rates which are statistically competitive with other SOTA systems. We also examine the contribution of individual agents and capabilities, and provide an error analysis to identify the strengths and weaknesses of our multi-agent approach, along with opportunities for improvement.
\end{enumerate}
\section{Related Work}

\paragraph{Single-Agent Approaches.} Recent advances in large language models (LLMs) such as GPT-4 \cite{openai2023gpt4} have renewed interest in the development of autonomous agents that can solve tasks on behalf of people \cite{russellnorvig1996, Wooldridge1995,Jennings1998Applications,wu2023autogen,yang2023autogptonlinedecisionmaking,sodhi2024stepstackedllmpolicies,zhang2024cognitivekernelopensourceagent,reed2022generalist}. These modern agents have shown remarkable skills in software development \cite{wang2024opendevinopenplatformai,zhang2024autocoderover,yang2024swe,xia2024agentless}, web manipulation \cite{deng2023mind2webgeneralistagentweb,zhang2024webpilotversatileautonomousmultiagent,zhou2024webarenarealisticwebenvironment,nakano2021webgpt,abuelsaad2024agenteautonomouswebnavigation}, manipulation of general graphical user interfaces \cite{zhang2024ufouifocusedagentwindows,wu2024oscopilotgeneralistcomputeragents,bonatti2024windowsagentarenaevaluating,pan2024autonomousevaluationrefinementdigital}, and other domains \cite{park2023generativeagentsinteractivesimulacra, Wang2023survey}. 
%Moreover, a subset of such agentic systems exhibit some generalist ability across different domains, however, they are not at a consistent state-of-the-art level across all domains \cite{wang2024opendevinopenplatformai,zhang2024cognitivekernelopensourceagent}.

Common strategies for developing such agents \cite{liu2024agent,xi2023risepotentiallargelanguage, masterman2024landscape, cheng2024exploring} include equipping LLMs with tools such as for code execution and web browsing \cite{qin2023tool,qin2023toolllm,schick-arxiv2023,mialon2023gaiabenchmarkgeneralai} and prompting strategies for better reasoning and planning such as CoT \cite{wei2022chain}, ReACT \cite{yao-iclr2023} and few-shot prompting \cite{zhou2024webarenarealisticwebenvironment}. With the development of multimodal models, agents can also operate in visual domains with techniques such as Set-of-Marks prompting \cite{yang2023set} among others \cite{yang2023set, zhang2024lookscreensmultimodalchainofaction,paranjape2023art,he2024webvoyagerbuildingendtoendweb}. To allow agents to accomplish tasks that require multiple steps with improved reliability, agent systems can incorporate self-critique \cite{wu2024oscopilotgeneralistcomputeragents,pan2024autonomousevaluationrefinementdigital,paul-etal-2024-refiner}, and inference-time search \cite{chen2024treesearchusefulllm,yao2023treethoughtsdeliberateproblem,koh2024treesearchlanguagemodel,song2024trialerrorexplorationbasedtrajectory}. Finally, Agentic systems can also benefit from memory and training either through explicit fine-tuning \cite{zeng2023agenttuningenablinggeneralizedagent,pan2024autonomousevaluationrefinementdigital,liu-arxiv2024,putta2024agentqadvancedreasoning} or through memory mechanisms \cite{wang2024agentworkflowmemory,sodhi2024stepstackedllmpolicies}.  Our work incorporates a subset of these techniques, and distributes them across agents in \team's multi-agent workflow, resulting in a modular, easy-to-extend implementation.

\paragraph{Multi-Agent Approaches.} The multi-agent paradigm presents an attractive modular and flexible approach to tackling complex tasks \cite{MASaSurvey2000, Messing2002AnIT, Grosz1999SharedPlans,Tambe1998AgentTeams, Scerri2001AdjustableAI, wu2023autogen, talebirad2023multiagentcollaborationharnessingpower,guo2024large,liu2024agent,xi2023risepotentiallargelanguage,masterman2024landscape,cheng2024exploring}. Commonly each agent either has access to different tools or has a different role in the team, sometimes defined through the system prompt of the LLM or by explicit training. Sibyl presents a multi-agent approach with a debate-based jury mechanism with tools for python code execution and web browsing \cite{wang2024sibylsimpleeffectiveagent}. WebPilot uses a multi-agent system with global and local optimization  in planning for web based tasks \cite{zhang2024webpilotversatileautonomousmultiagent}. Trase claims to use a multi-agent architecture with a top level agent with self-critique and lower level agents \cite{redcell_trase_2024}.  A host of other multi-agent systems and frameworks have also been introduced \cite{li2023camel,liang-arxiv2023,du2023improving,babyagi,hong2023metagpt}. However, the previous methods differ from the architecture of \team which incorporates dynamic routing between agents using the Orchestrator along with planning and recovery.

\paragraph{Agentic Evaluation.} To evaluate agents on general multi-step tasks, numerous benchmarks have been proposed in the literature \cite{mialon-arxiv2023,zhou2024webarenarealisticwebenvironment,xie2024osworldbenchmarkingmultimodalagents,liu2024visualwebbenchfarmultimodalllms,yoran2024assistantbenchwebagentssolve,yao2023webshopscalablerealworldweb,shi2017world,deng2023mind2webgeneralistagentweb,pan2024webcanvasbenchmarkingwebagents,li2024websuite}. Given the general and ubiquitous nature of the web, many of these benchmarks heavily incorporate \cite{mialon2023gaiabenchmarkgeneralai, yoran2024assistantbenchwebagentssolve}, or exclusively consider \cite{zhou2024webarenarealisticwebenvironment, deng2023mind2webgeneralistagentweb} browser-based tasks. These benchmarks either rely on non-interactive traces through real websites such as Mind2Web \cite{deng2023mind2webgeneralistagentweb}, interaction with synthetically created websites such as in WebArena \cite{zhou2024webarenarealisticwebenvironment}, or interaction with real websites on the public Internet such as GAIA \cite{mialon2023gaiabenchmarkgeneralai}. In the former case, non-interactive benchmarks are  limiting for evaluating agentic systems since they do not allow agents to deviate from previously recorded paths. This makes it impossible to evaluate error recovery, or find novel alternative strategies for the given problem.
Therefore, we focus on benchmarks that rely on interacting with live websites -- whether synthetic or public -- as they are more faithful to real-world tasks. Moreover, we prioritize benchmarks such as GAIA, which test generalist skills like data analysis or coding, in addition to commanding web browsers to navigate pages. We contribute \agbench{} as a standalone tool to perform evaluation of agentic systems, relying on benchmarks from the literature. Furthermore, we provide an in-depth error analysis of \team's performance contributing to work on debugging agentic systems \cite{li2024websuite}.

\section{Problem Setup}

\paragraph{Complex Tasks.} In this work our goal is to build a generalist agentic system capable of solving \emph{complex tasks} across a variety of domains. We define a task as {\em complex} if it requires, or significantly benefits from, a process involving planning, acting, observing, and reflecting, potentially multiple times. Acting refers to more than generating tokens, such as executing code, using tools, or interacting in an environment. Observing, in this context, provides information that was previously unavailable or unknowable. A task is defined by an input, a desired output and an evaluation function to compare the desired output to any candidate output. The input consists of a well-specified textual description and an optional arbitrary set of file attachments which may include images, dataset files, audio clips among other things. For example, the input task description could be \textit{``fact-check each claim in the attached PDF as correct or incorrect"} with a PDF file as an attachment. The desired output consists either of a textual answer (possibly representing a structured object), or a specific state of the environment to reach. In the fact-checking example, the output might be a string labeling each fact as correct or not, e.g., \textit{``claim 1: correct, claim 2: incorrect, ..."}.
Here, the evaluation function might simply determine whether the desired output and the proposed answer match exactly.

\paragraph{Agentic Systems.} 
To complete a task, assume a {\em computer} which can be partially observed and operated to complete the task. The computer constitutes the \emph{environment}.
An agentic system can take as input the task description, and any related attachments that are present on the computer environment. The system is allowed to do arbitrary processing to complete the task, but must complete it within a time budget (e.g., 25 mins).
For instance, on the computer, the autonomous system can execute Python code, navigate the web using a browser, download files locally, among other actions from its \emph{action space}. The system's ability to take action in, and potentially modify, both the local and web environments is why we refer to the system as \emph{agentic}. After completing the task, the system returns a text answer, and a trace of its observations and steps along the way. The final state of the environment is also captured in sufficient detail to run the task evaluation. 
Note that this setting can be described as a Partially Observable Markov Decision Process, similar to formalizations used by prior work~\cite{sodhi2024stepstackedllmpolicies}.
Nex, we describe \team, our multi-agent system that can autonomously solve complex tasks. 
\section{\team Overview}

\begin{figure}[!t]
    \centering
    \includegraphics[width=\linewidth]{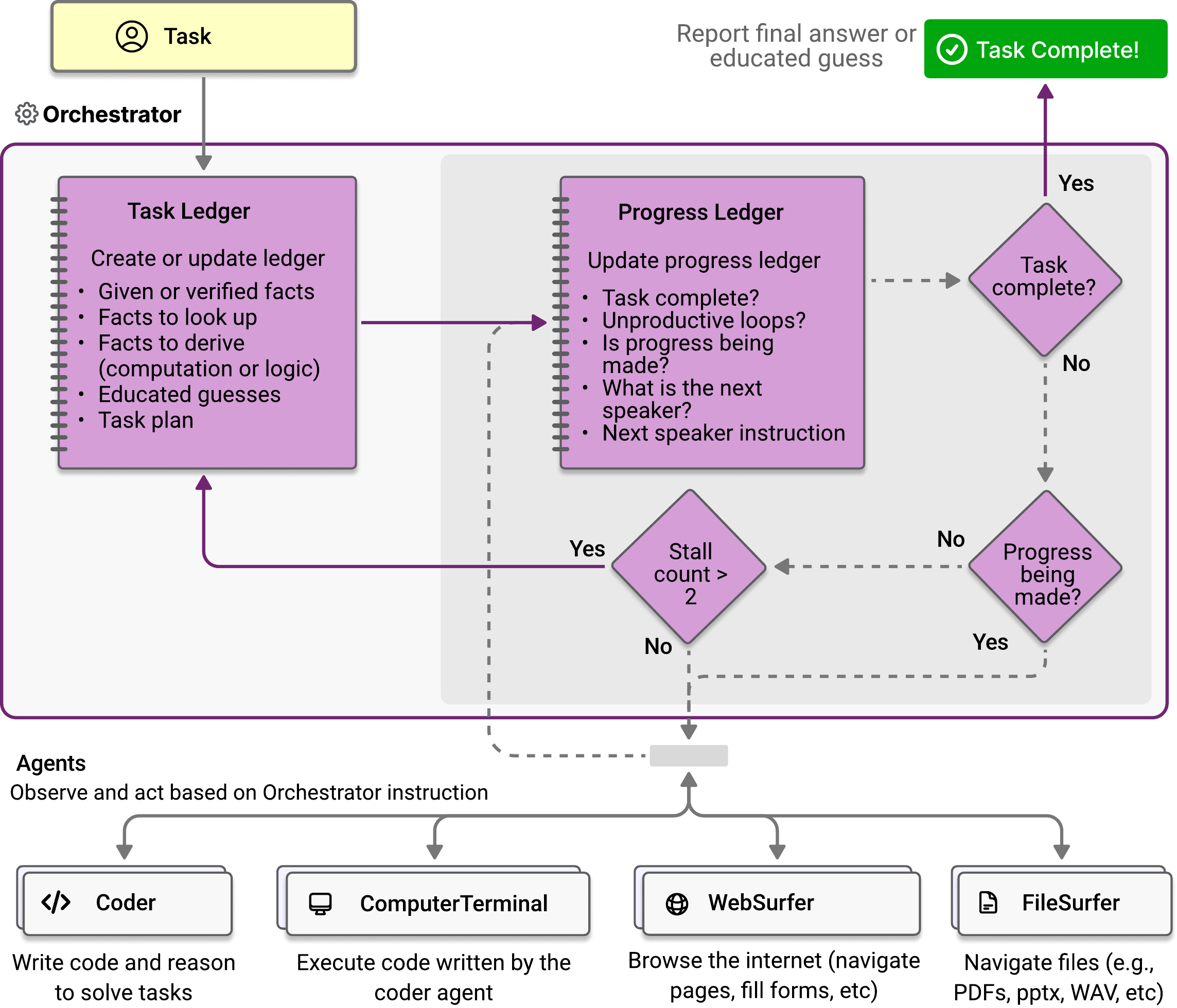}
    \caption{\team{} features an Orchestrator agent that implements two loops: an outer loop and an inner loop. The outer loop (lighter background with solid arrows) manages the task ledger (containing facts, guesses, and plan). The inner loop (darker background with dotted arrows) manages the progress ledger (containing current progress, task assignment to agents).}
    \label{fig:magentic-orchestrator}
\end{figure}

\team is a generalist multi-agent system for autonomously completing complex tasks. The team's work is coordinated by an Orchestrator agent, responsible for task decomposition and planning, directing other agents in executing subtasks, tracking overall progress, and taking corrective actions as needed. The other agents on the team are specialized with different capabilities necessary for completing ad-hoc, open-ended tasks such as browsing the web and interacting with web-based applications, handling files, and writing and executing Python code (Figure~\ref{fig:magentic-orchestrator}). 

Together, the \team team collaborates to solve tasks on behalf of a user. For example, suppose a user requests a survey and concise slide presentation of AI safety papers published in the last month. \team will approach this task as follows. The Orchestrator will first create a plan that breaks down the task into subtasks, such as searching for abstracts, downloading relevant papers, reading and summarizing the papers, and finally creating a presentation out of the findings. This initial plan serves as providing a guide or rubric for acting, and may not be followed exactly. Instead it can be interpreted as similar to chain of thought prompting for the agents \cite{wei2022chain}. Once this initial plan is formed, the Orchestrator then selects an appropriate agent and assigns it a subtask. For example, the WebSurfer agent might be directed to search for and download AI safety papers, while the FileSurfer agent might be directed to open the downloaded PDFs and extract relevant information. The Coder agent might create the presentation by writing Python code to interact with various files, and the ComputerTerminal agent would then execute the code written to produce the final output (or to report execution errors the coder agent has yet to address). As the task progresses, the Orchestrator coordinates the agents, monitors progress, and monitors for task completion. 

In the following sections, we first describe \team's inter-agent workflow, driven by the Orchestrator, then describe each individual agent's design, capabilities, and action space.

\subsection{\team's Multi-Agent Workflow}
Figure \ref{fig:magentic-orchestrator} illustrates \team's workflow in more depth. At a high level, the workflow contains two loops, the outer loop maintains the {\em task ledger}, which contains the overall plan, while  the inner loop maintains the {\em progress ledger}, which directs and evaluates the individual steps that contain instructions to the specialized agents. 

\paragraph{Outer Loop.} The outer loop is triggered by an initial prompt or task. In response, the Orchestrator creates the task ledger to serve as short-term memory for the duration of the task. Upon receiving the task, the Orchestrator reflects on the request and pre-populates the task ledger with vital information-- given or verified facts, facts to look up (e.g., via web search), facts to derive (e.g., programmatically, or via reasoning), and educated guesses. These initial educated guesses are important, and can allow the Orchestrator to express memorized closed-book information in a guarded or qualified manner, allowing agents to potentially benefit, while lessening the system's overall sensitivity to errors or hallucinations. For example, agents might only rely on the guesses when they get stuck, or when they run out of time and need to output a best guess for the benchmark. Educated guesses are updated periodically, by the outer loop, as new information becomes available.

Only after the facts and guesses in the task ledger have been populated, the Orchestrator considers the makeup of the team it is directing. Specifically, it uses each team member's description, along with the current task ledger, to devise a step-by-step plan. The plan is expressed in natural language and consists of a sequence of steps and assignments of those steps to individual agents. Since the plan is used in a manner similar to chain of thought prompting \cite{wei2022chain}, it serves more as a hint for step-by-step execution -- neither the Orchestrator nor the other agents are required to follow it exactly.  Since this plan may be revisited with each iteration of the outer loop, we force all agents to clear their contexts and reset their states after each plan update. Once the plan is formed, the inner loop is initiated.

\paragraph{Inner Loop.} During each iteration of the inner loop, the Orchestrator answers five questions to create the progress ledger:

{\em 
\begin{itemize}
\setlength\itemsep{0em}
\item Is the request fully satisfied (i.e., task complete)?
\item Is the team looping or repeating itself?
\item Is forward progress being made?
\item Which agent should speak next?
\item What instruction or question should be asked of this team member?
\end{itemize}
}

\noindent 
When answering these questions, the Orchestrator considers both the task ledger (containing facts, guesses, and a plan), and the current agent conversation context. 

The Orchestrator also maintains a counter for how long the team has been stuck or stalled. If a loop is detected, or there is a lack of forward progress, the counter is incremented. As long as this counter remains below a threshold ($\leq 2$ in our experiments), the Orchestrator initiates the next team action by selecting the next agent and its instruction.  However, if the counter exceeds the threshold, 
the Orchestrator breaks from the inner loop, and proceeds with another iteration of the outer loop. This includes initiating a reflection and self-refinement step \cite{shinn2024reflexion}, where it identifies what may have gone wrong, what new information it learned along the way, and what it might do differently on the next iteration of the outer loop. It then updates the task ledger, revises the original plan, and starts the next cycle of inner loop. Together, this counter-based mechanism gives the agents a limited budget to recover from small errors, or to persist through brief episodes of uncertainty in progress.

This nested-loop behavior continues until the Orchestrator determines the task is complete or the team has reached some (parameterized and configurable) termination logic, such as reaching a maximum number of attempts, or exceeding a specified maximum time limit. 

Finally, upon termination of both loops, the Orchestrator reviews the full transcript, along with the ledger, and reports either a final answer, or its best educated guess.

\subsection{\team's Agents}

The Orchestrator agent in \team coordinates with four specialized agents: WebSurfer, FileSurfer, Coder and ComputerTerminal. As the names suggest, each of these agents is optimized for a specific -- yet generally useful -- capability. In most cases, these agents are constructed around LLMs with custom system prompts, and capability-specific tools or actions. For example, WebSurfer can navigate to pages, click links, scroll the viewport, etc. In other cases, agents may operate deterministically, and do not include LLMs calls at all. For example, the ComputerTerminal deterministically runs Python code, or shell commands, when asked. 

This decomposition of high-level capabilities \emph{across} agents, and low-level actions \emph{within} agents, creates a hierarchy over tool usage which may be easier for the LLMs to reason about. For example, rather than deciding between dozens of possible actions, the Orchestrator needs only to decide which agent to call to access a broad capability (e.g., browsing the web). The chosen agent then selects from a limited set of agent-specific actions (e.g., clicking a button versus scrolling the page). 

We detail the implementation of each of the agents below:

\begin{itemize}
    \item \textbf{WebSurfer}:
    This is a highly specialized LLM-based agent that is proficient in commanding and managing the state of a Chromium-based web browser. With each incoming natural-language request, the WebSurfer maps the request to a single action in its action space (described below), then reports on the new state of the web page (providing both a screenshot and a written description). As an analogy, this configuration resembles a telephone technical support scenario where the Orchestrator knows what to do, but cannot directly act on the web page. Instead it relays instructions, and relies on the WebSurfer to carry out actions and report observations. 
    
    The action space of the WebSurfer includes navigation (e.g. visiting a URL, performing a web search, or scrolling within a web page);  web page actions (e.g., clicking and typing); and reading actions (e.g., summarizing or answering questions). This latter category of reading actions allows the WebSurfer to directly employ document Q\&A techniques in the context of the full document. This saves considerable return-trips to the orchestrator (e.g., where the orchestrator might simply command the agent to continue scrolling down), and is advantageous for many tasks. 
    
    When interacting with web page elements (e.g.,  when clicking or typing), the WebSurfer must ground the actions to specific coordinates or elements of the current web page. For this we use set-of-marks prompting \cite{yang2023set} in a manner similar to Web Voyager\cite{he2024webvoyagerbuildingendtoendweb}. This step relies on an annotated screenshot of the page, and thus is inherently multi-modal. We further extended the set-of-marks prompt to include textual descriptions of content found {\em outside} the visible view port, so that the agent can determine what might be found by scrolling \footnote{Scrolling is needed because, like human users, the WebSurfer agent cannot interact with page elements that are outside the active viewport.}, or opening menus or drop-downs.
    
    \item \textbf{FileSurfer}:
    The FileSurfer agent is very similar to the WebSurfer, except that it commands a custom markdown-based file preview application rather than a web browser. This file preview application is read-only, but supports a wide variety of file types, including PDFs, Office documents, images, videos, audio, etc.  The FileSurfer can also perform common navigation tasks such as listing the contents of directories, and navigating a folder structure.

    \item \textbf{Coder}: This is an LLM-based agent specialized through its system prompt for writing code, analyzing information collected from the other agents, or creating new artifacts. The coder agent can both author new programs and debug its previous programs when presented with console output.   
    
    \item \textbf{ComputerTerminal}: Finally, the ComputerTerminal provides the team with access to a console shell where the Coder's programs can be executed. ComputerTerminal can also run shell commands, such as to download and install new programming libraries. This allows the team to expand the available programming tool set, as needed. 
\end{itemize}

Together, \team's agents provide the Orchestrator with the tools and capabilities that it needs to solve a broad variety of open-ended problems, as well as the ability to autonomously adapt to, and act in, dynamic and ever-changing web and file-system environments.

\section{Experiments}
\label{sec:Experiments}

\subsection{\agbench \xspace and Setup}
\paragraph{Overview.} Agentic systems, such as \team, that interact with stateful environments, pose unique challenges for evaluation. For example, if a task requires installing a Python library, the first system to be evaluated will be disadvantaged: Its agents will have to first write Python code that fails, then debug the problem, install the library, and finally try again. Subsequent runs -- perhaps with other agents or models -- will then benefit from the library's presence, and thus may appear to perform better simply because they were executed later. Conversely, an erroneous agent could take actions (e.g. deleting files, or placing the the system in an inoperable state), that would harm all future tasks. To this end, it is crucial that any evaluation be independent across tasks, and provide safety from dangerous side effects (e.g., from agents' actions).

To address this challenge, we developed \agbench for evaluating agentic systems. Given a benchmark, which consists of a set of independent tasks and associated evaluation functions, \agbench{} allows users to run these tasks in a setting with tightly controlled initial conditions: in each task, \agbench{} will start from a blank slate with freshly initialized Docker containers, providing the recommended level of consistency and safety.
The results of each task are logged in a central location on the host machine (outside of Docker), and can be ingested for analysis by metrics scripts. Furthermore, \agbench{} allows users to launch multiple tasks in parallel to speed up evaluation, or to compute variance across repeated runs. 

\paragraph{Benchmarks.} Using \agbench, we can implement and evaluate \team on a variety of benchmarks. Our criteria for selecting benchmarks is that they should involve complex multi-step tasks, with at least some tasks or steps requiring planning and tool use ( including using web browsers to act on real or simulated webpages, handling files, etc.) We consider three benchmarks in this work that satisfy this criteria: GAIA, AssistantBench, and WebArena.

\emph{GAIA} \cite{mialon2023gaiabenchmarkgeneralai} is a benchmark for general AI assistants with 465 multi-modal question--answer pairs that are real-world and challenging, requiring multiple steps and multiple tools to solve (e.g., navigating the web, handling files, etc.). Despite the complexity of the tasks, GAIA questions are designed to be automatically and unambiguously verifiable, with each answer consisting of a target string that can be checked by string matching. GAIA is split into an open validation set with 165 question--answer pairs, and a test set with 300 questions (answers hidden).\footnote{Leaderboard: \url{https://gaia-benchmark-leaderboard.hf.space/}}
An example of a GAIA task follows:
\begin{quote}
\textbf{Example GAIA task:} Of the cities within the United States where U.S.\ presidents were born, which two are the farthest apart from the westernmost to the easternmost going east, giving the city names only? Give them to me in alphabetical order, in a comma-separated list.
\end{quote}
In order to solve this task, one needs to perform multiple steps: use the web to find the birth city of each U.S.\ president, retrieve the coordinates of these cities, identify the westernmost and easternmost coordinates, then return the corresponding cities in alphabetical order. This requires web navigation, coding, and reasoning abilities, illustrating the complexity of GAIA. 

The second benchmark we consider is \emph{AssistantBench} \cite{yoran2024assistantbenchwebagentssolve}. Similar in design to GAIA, \emph{AssistantBench} is a set of 214 question--answer pairs that are realistic, time-consuming (requiring a human several minutes to perform), and automatically verifiable. They require navigating real-world websites and multi-step reasoning. As with GAIA, answers are evaluated by string matching, but AssistantBench introduces an additional softer metric of accuracy that affords a degree of partial credit \cite{yoran2024assistantbenchwebagentssolve}. AssistantBench is split into an open validation set with 33 question--answer pairs and a test set with 181 questions (answers hidden).%
\footnote{Leaderboard: \url{https://huggingface.co/spaces/AssistantBench/leaderboard}}
An example of an AssistantBench task follows:
\begin{quote}
\textbf{Example AssistantBench task:} Which supermarkets within 2 blocks of Lincoln Park in Chicago have ready-to-eat salad for under \$15?
\end{quote}
This task requires the agent to use an online map (e.g., Bing Maps) to find supermarkets near Lincoln Park, and then, for each supermarket found, to navigate to its website and check if it has ready-to-eat salads under \$15. 

The final benchmark we consider is \emph{WebArena} \cite{zhou2024webarenarealisticwebenvironment}, which involves performing complex tasks in a synthetic web environment. Each task requires multi-step planning and acting, and targets one or more fully functional synthetic websites. It contains 812 tasks across five major website categories (e.g., shopping, forums, maps, etc.), and a sixth category that requires interacting with multiple websites.
Tasks are evaluated by running per-task evaluation scripts in the context of the running website to check that answers exactly or approximately match a target, and that the page is left in the desired state (e.g., that a comment has been posted, or an item is in a shopping cart). There is a public leaderboard for WebArena, but it is based on self-reported results.
\footnote{Leaderboard: \url{https://docs.google.com/spreadsheets/d/1M801lEpBbKSNwP-vDBkC_pF7LdyGU1f_ufZb_NWNBZQ/edit}}
The dataset also provides no formal validation / test split across tasks \cite{kapoor2024aiagentsmatter}. We developed our own split so that we might assess \team{}'s ability to generalize to tasks in the unseen test set --  which was evaluated only once. To split the tasks, we computed the MD5 hash of each problem's \emph{template\_id}\footnote{WebArena tasks are populated by expanding a smaller number of task templates.}, then assigned the 422 tasks with hashes starting with digits 0-7 to the validation set (the remaining 390 tasks were assigned to the test set). An example of a WebArena task, from the validation set, is as follows:

\begin{quote}
\textbf{Example WebArena task:} Tell me the count of comments that have received more downvotes than upvotes for the user who made the latest post on the Showerthoughts forum.
\end{quote}
To solve this task, the agents have to navigate the Showerthoughts forum, find the profile of the user with the latest post, retrieve all their comments, and finally count those with more downvotes than upvotes. This illustrates the multi-step navigation nature of WebArena tasks.

\paragraph{Implementation Details.} 
An identical configuration of \team was used for all three benchmarks, but some additional set up code was needed for each. Namely, each benchmark used a unique final prompt to ensure answers were expressed in the benchmark-specific prescribed format. Additionally, set up code for WebArena included instructions to log in to websites, which is not considered part of the task. Finally, WebArena refers to the Postmill website as Reddit,\footnote{WebArena's Postmill website is populated from data crawled from Reddit}, causing agents to complain that they were on the wrong website. To address this, we included the following prompt text:

``\emph{[This website is] a Postmill forum populated with a large sample of data crawled from Reddit. Postmill is similar to Reddit, but the UI is distinct, and 'subreddits' begin with /f/ rather than /r/}``

We include similar prompts for the three other WebArena sites, and we discuss this issue more in section \ref{sec:Risks and Mitigations}.

For \team, the default multimodal LLM we use for all agents (except the ComputerTerminal) is \textit{gpt-4o-2024-05-13}. In a different configuration of \team, we experiment with using OpenAI o1-preview\footnote{\url{https://openai.com/index/introducing-openai-o1-preview/}} for the outer loop of the Orchestrator and for the Coder, while other agents continue to use GPT-4o. In this case, only a subset of the agents (e.g., the WebSurfer) are multimodal since o1-preview can process only text as input. We implement \team on the multi-agent platform AutoGen version 0.4 \cite{wu2023autogen}. The code for \team is made publicly available.\footnote{\url{https://aka.ms/magentic-one}} The experiments reported here were conducted between August and October 2024. 

\subsection{Results}
\label{sec:Results}

\paragraph{Results.} Table \ref{tab:all_results} shows the performance of \team  compared to relevant baselines for all three benchmarks. For GAIA and AssistantBench, we report only results for the test sets. For WebArena there is no common test set, so we report results for all 812 tasks. We separately show performance of \team when using only GPT-4o as the model for all agents, and when using a combination of GPT-4o and o1-preview.\footnote{We do not report results for \team (GPT-4o, o1) on WebArena since the o1 model refused to complete 26\% of WebArena Gitlab tasks, and 12\% of Shopping Administration tasks, making a fair comparison impossible.} We also include the highest-performing baselines in the literature, for each benchmark, according to the leaderboards as of October 21, 2024. This includes entries that are neither open-source, nor described by technical reports, making them difficult to independently validate. Finally, we also include human performance where available.

We use statistical tests to compare the performance of \team to baselines and say that two methods are statistically comparable if the difference in their performance is not statistically significant ($\alpha$=0.05); details about our statistical methodology can be found in Appendix \ref{apx:stats_results}.

\team (GPT-4o, o1-preview) achieves statistically comparable performance to SOTA methods on both GAIA and AssistantBench. On WebArena, only the GPT-4o variant was evaluated\footnotemark[12], and it achieved comparable performance to most SOTA methods except for WebPilot \cite{zhang2024webpilotversatileautonomousmultiagent} and Jace.AI (which achieve statistically higher scores).

As noted earlier, WebArena does not have a hidden test set, thus posing some awkward challenges for fair evaluation. To investigate this, we consider the self-imposed validation/test splits that we created apriori. On the the validation set, \team{} correctly performed 35.1\% of tasks (148 of 422), falling to 30.5\% (119 of 390) for the test set. When setting up the WebArena benchmark, we used the validation set to initially validate and debug our workflow. This result suggests that extra attention paid on validation tasks has lead to at least mild over-fitting. It is unclear whether other entries on the leaderboard performed similar analyses or took similar precautions. We would encourage the WebArena authors to develop a hidden test set for future comparison purposes. 

Comparing \team (GPT-4o) and \team (GPT-4o, o1), the biggest gains are observed on the GAIA benchmark. We hypothesize that this occurs because GAIA involves tasks that require more logical reasoning and puzzle-solving compared to AssistantBench. These are skills for which o1 was optimized.

Together, these results establish \team as a strong agentic system for completing complex web- and file-based tasks. Moreover, achieving this level of performance across benchmarks speaks to the team's generality -- note that among the baselines in Table \ref{tab:all_results}, no prior system (other than base models) has been been evaluated across all three benchmarks. 

\begin{table}[h!]
\centering
\caption{Performance of \team compared to relevant baselines on the test sets of GAIA, WebArena and AssistantBench. For each method we note in parenthesis the LLM used to obtain the result. The numbers reported denote exact task completion rate as a percentage. All results for baselines are obtained from the corresponding benchmark leaderboard. We do not report results for \team (GPT-4o, o1) on WebArena since the o1 model refused to complete 26\% of WebArena Gitlab tasks, and 12\% of Shopping Administration tasks, making a fair comparison impossible. An example task refused by o1 is ``\emph{create a new group "webagent" with members pandey2000, sayakpaul, sayakpaul}``. We include 95\% error bars as $\pm$ using the Wald interval method. We underline results that are statistically comparable to \team (GPT-4o, o1) according to a z-test with $\alpha=0.05$, and bold results that statistically exceed our performance (Appendix \ref{apx:stats_results}). 
}

\resizebox{1\textwidth}{!}{
\begin{tabular}{p{0.32\textwidth}p{0.11\textwidth}p{0.15\textwidth}p{0.15\textwidth}p{0.12\textwidth}}
\toprule
Method & GAIA & AssistantBench (EM) & AssistantBench (accuracy) & WebArena \\
\midrule
omne v0.1 (GPT-4o, o1)  & \underline{40.53$\pm$5.6} & -- & -- & -- \\
Trase Agent v0.2 (GPT-4o, o1, Gemini) & \underline{39.53$\pm$5.5}& -- & -- & -- \\
Multi Agent (NA) & \underline{38.87$\pm$5.5}& -- & -- & -- \\
das agent v0.4 (GPT-4o) & \underline{38.21$\pm$5.5}& -- & -- & -- \\
Sibyl (GPT-4o) \cite{wang2024sibylsimpleeffectiveagent} & \underline{34.55$\pm$5.4} & -- & -- & -- \\
HF Agents (GPT-4o)  & \underline{33.33$\pm$5.3}& -- & -- & -- \\
FRIDAY (GPT-4T) \cite{wu2024oscopilotgeneralistcomputeragents} & 24.25$\pm$4.8 & -- & -- & -- \\
GPT-4 + plugins \cite{mialon2023gaiabenchmarkgeneralai}  & 14.60$\pm$4.0 & -- & -- & --\\
SPA $\rightarrow$ CB (Claude) \cite{yoran2024assistantbenchwebagentssolve}  & -- & \underline{13.8$\pm$5.0} & \underline{26.4$\pm$6.4} & -- \\
SPA $\rightarrow$ CB (GPT-4T) \cite{yoran2024assistantbenchwebagentssolve}  & -- &\underline{ 9.9$\pm$4.3 }& \underline{25.2$\pm$6.3} & -- \\
Infogent (GPT-4o) & -- & 5.5$\pm$3.3 & 14.5$\pm$5.1 & -- \\
Jace.AI (NA)  & -- & -- & -- & \textbf{57.1$\pm$3.4} \\
WebPilot (GPT-4o)  \cite{zhang2024webpilotversatileautonomousmultiagent} & -- & -- & -- &\textbf{37.2$\pm$3.3} \\
AWM (GPT-4) \cite{wang2024agentworkflowmemory} & -- & -- & -- & \underline{35.5$\pm$3.3}\\
SteP (GPT-4) \cite{sodhi2024stepstackedllmpolicies} & -- & -- & -- & \underline{33.5$\pm$3.2} \\
BrowserGym  (GPT-4o) \cite{drouin2024workarenacapablewebagents} & -- & -- & -- & 23.5$\pm$2.9 \\

GPT-4  & 6.67$\pm$2.8\cite{mialon2023gaiabenchmarkgeneralai} & 6.1 $\pm$3.5\cite{yoran2024assistantbenchwebagentssolve} & 16.5 $\pm$5.4\cite{yoran2024assistantbenchwebagentssolve} & 14.9$\pm$2.4\cite{zhou2024webarenarealisticwebenvironment} \\
\midrule
Human & 92.00$\pm$3.1 & -- & -- & 78.2$\pm$2.8 \\
\midrule
\textbf{\team}  (GPT-4o) & 32.33$\pm$5.3& \underline{11.0 $\pm$4.6}& \underline{25.3 $\pm$6.3}& \underline{32.8$\pm$3.2}\\
\textbf{\team}  (GPT-4o, o1) &\underline{38.00$\pm$5.5}& \underline{13.3 $\pm$4.9}& \underline{27.7 $\pm$6.5}& * \\
\bottomrule
\end{tabular}
}
\label{tab:all_results}

\end{table}

\paragraph{Performance Breakdown by Task Difficulty or Domain} Each benchmark provides some categorization of tasks by difficulty (GAIA, AssistantBench), or application domain (WebArena). In Table \ref{tab:results_by_category}, we breakdown performance by category, comparing \team{}to the best-performing baselines for GAIA and AssistantBench, and to WebPilot \cite{zhang2024webpilotversatileautonomousmultiagent}, the best performing WebArena baseline for which category-level results are available.  

By breaking down performance by category, we immediately notice that \team appears to compete better on hard tasks (e.g., level 3, hard) vs. easy tasks (e.g. level 1, easy). In fact, on AssistantBench, \team outperforms the best comparable baseline on the hardest category. Similarly, on WebArena, \team differs from WebPilot mainly on the Reddit category -- again the apparent easiest category by score.

We hypothesize that \team introduces some fixed overhead or complexity that disproportionately helps with long multi-step tasks, while introducing more opportunities for errors on short few-step tasks. This presents an opportunity to enhance \team for simpler tasks to achieve SOTA across all levels. 

\begin{table}[h]
\centering
\caption{Performance comparison between \team (GPT-4o), \team (GPT-4o, o1) and the best baseline for each benchmark's test set. Analysis is split across the different categories of each benchmark. Since there is no available baseline that evaluates on all three benchmarks, we picked the best baseline with available results per benchmark. The best baseline for GAIA is omne v0.1. The best baseline for WebArena with available category wise results is WebPilot \cite{zhang2024webpilotversatileautonomousmultiagent}. The best baseline for AssistantBench is SPA $\rightarrow$ CB (Claude) \cite{yoran2024assistantbenchwebagentssolve}. For WebArena, top leaderboard methods \cite{sodhi2024stepstackedllmpolicies,zhang2024webpilotversatileautonomousmultiagent} consider the cross site tasks in WebArena as belonging to one of the 5 sites, and so the comparison with \team{} may  differ. } 
\resizebox{1\textwidth}{!}{
\begin{tabular}{p{0.2\textwidth}p{0.15\textwidth}p{0.2\textwidth}p{0.20\textwidth}p{0.20\textwidth}}
\toprule
\textbf{Dataset} & \textbf{Category} & \textbf{\team} (GPT-4o) & \textbf{\team} (GPT-4o, o1) & Best Baseline \cite{zhang2024webpilotversatileautonomousmultiagent,yoran2024assistantbenchwebagentssolve}  \\
\midrule
\multirow{4}{*}{GAIA \cite{mialon2023gaiabenchmarkgeneralai}} 
 & Level 1 & 46.24 &  \textbf{54.84} & 53.76  \\
 & Level 2 & 28.3 & 32.7 & \textbf{37.11} \\
 & Level 3 & 18.75 &  22.92 & \textbf{26.53} \\

 \midrule
 \multirow{4}{*}{AssistantBench \cite{yoran2024assistantbenchwebagentssolve}} 
 & Easy & 69.9 &  73.4 & \textbf{81}  \\
 & Medium & 35.6 &  \textbf{47.1} & 44.6 \\
 & Hard & \textbf{16.9} &  14.8 & 13.3 \\
 \midrule
\multirow{7}{*}{WebArena \cite{zhou2024webarenarealisticwebenvironment}} 
 & Reddit & 53.77 &  -- & \textbf{65.1} \\
 & Shopping & 33.16 &  -- & \textbf{36.}9 \\
 & CMS & \textbf{29.1} &  -- & 24.7 \\
 & Gitlab & 27.78 &  -- & \textbf{39.4} \\
 & Maps & \textbf{34.86} &  -- & 33.9\\
 & Cross Site & 14.6 &  -- & -- \\
\bottomrule
\end{tabular}
}
\label{tab:results_by_category}
\end{table}

\subsection{Ablations}
In this section, we examine how different agents and capabilities contribute to \team's performance through ablation experiments.

\paragraph{Setup.} On the validation set of GAIA \cite{mialon2023gaiabenchmarkgeneralai}, we perform multiple ablation experiments to evaluate the impact of key \team (GPT-4o) agents and components. First, to understand the impact of \team's Orchestrator, the AutoGen\cite{wu2023autogen} library's GroupChat mechanism. This baseline orchestrator simply decides which agent should speak next during task execution, eliminating both ledgers, planning, progress tracking, loop detection, and explicit instructions to other agents. The second set of ablations we perform is to remove individual agents from the \team team to measure the impact of those agents on overall task performance.

For all ablations, we report on results broken down by difficulty level and \emph{capabilities} required. For the capabilities analysis, we mapped the tools needed to complete tasks, as reported by human annotators of the GAIA dataset \cite{mialon2023gaiabenchmarkgeneralai}, to four categories: web browsing, coding, file handling, and none. These categories roughly correspond to the categories defined in \cite{mialon2023gaiabenchmarkgeneralai}, with minor adjustments to better align to the core functional-responsibilities of \team's agents. For example, the original categories in \cite{mialon2023gaiabenchmarkgeneralai} included a multi-modality category since multi-modal task handling was accomplished via a tool. However, because \team leverages multi-modal models, multi-modality is handled inherently by all agents rather than through use of a specific tool. In such cases, we noted the task as requiring no tools (i.e., 'none') to complete. Our capability mapping is described further in Appendix \ref{apx:capability_mapping}.

\begin{figure}[h!]
  \centering
  \begin{subfigure}[b]{0.8\textwidth}
    \includegraphics[width=\linewidth]{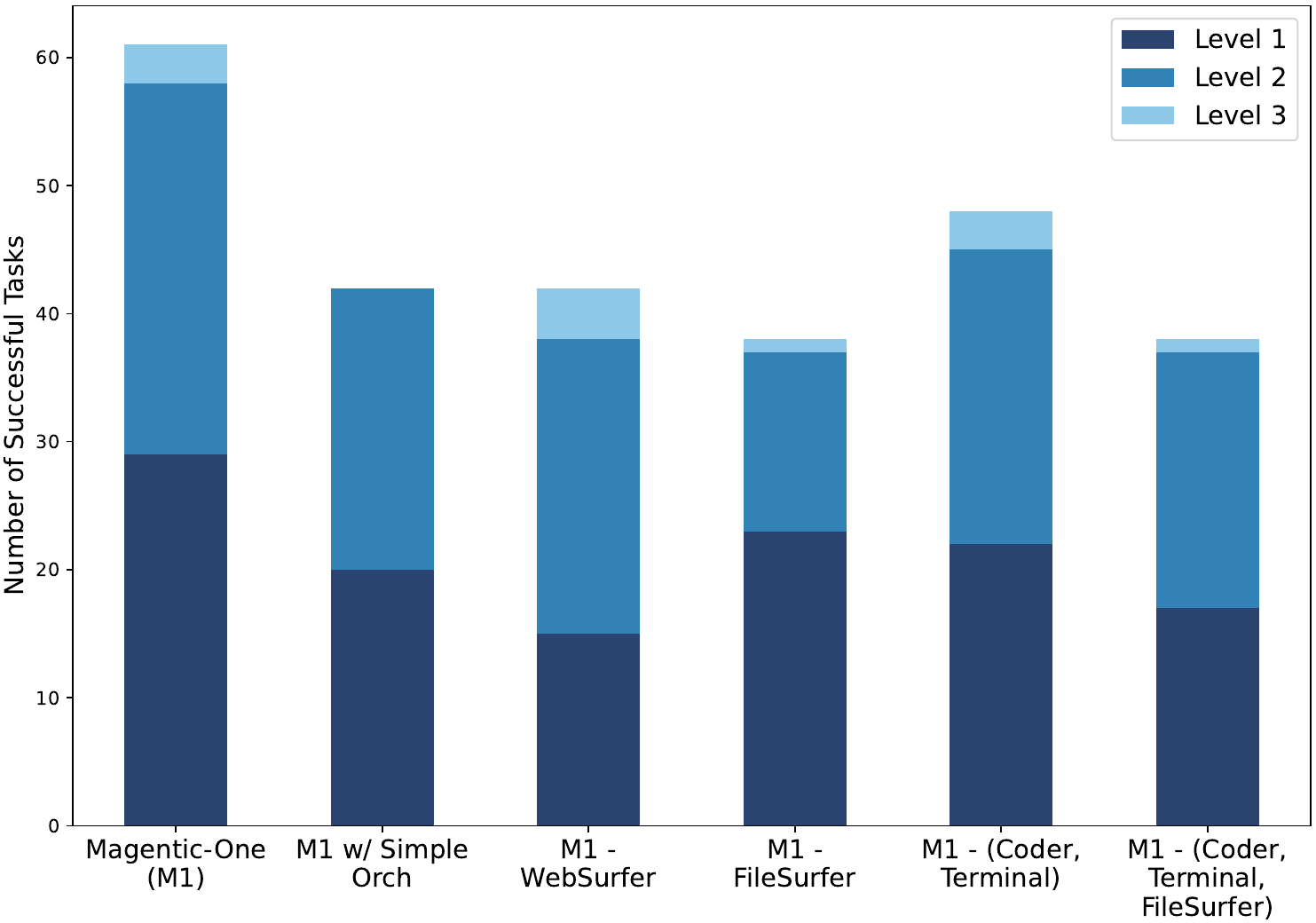}
    \caption{Performance by Level}
    \label{fig:ablations_level}
  \end{subfigure}
  \hfill
  \begin{subfigure}[b]{0.8\textwidth}
    \includegraphics[width=\linewidth]{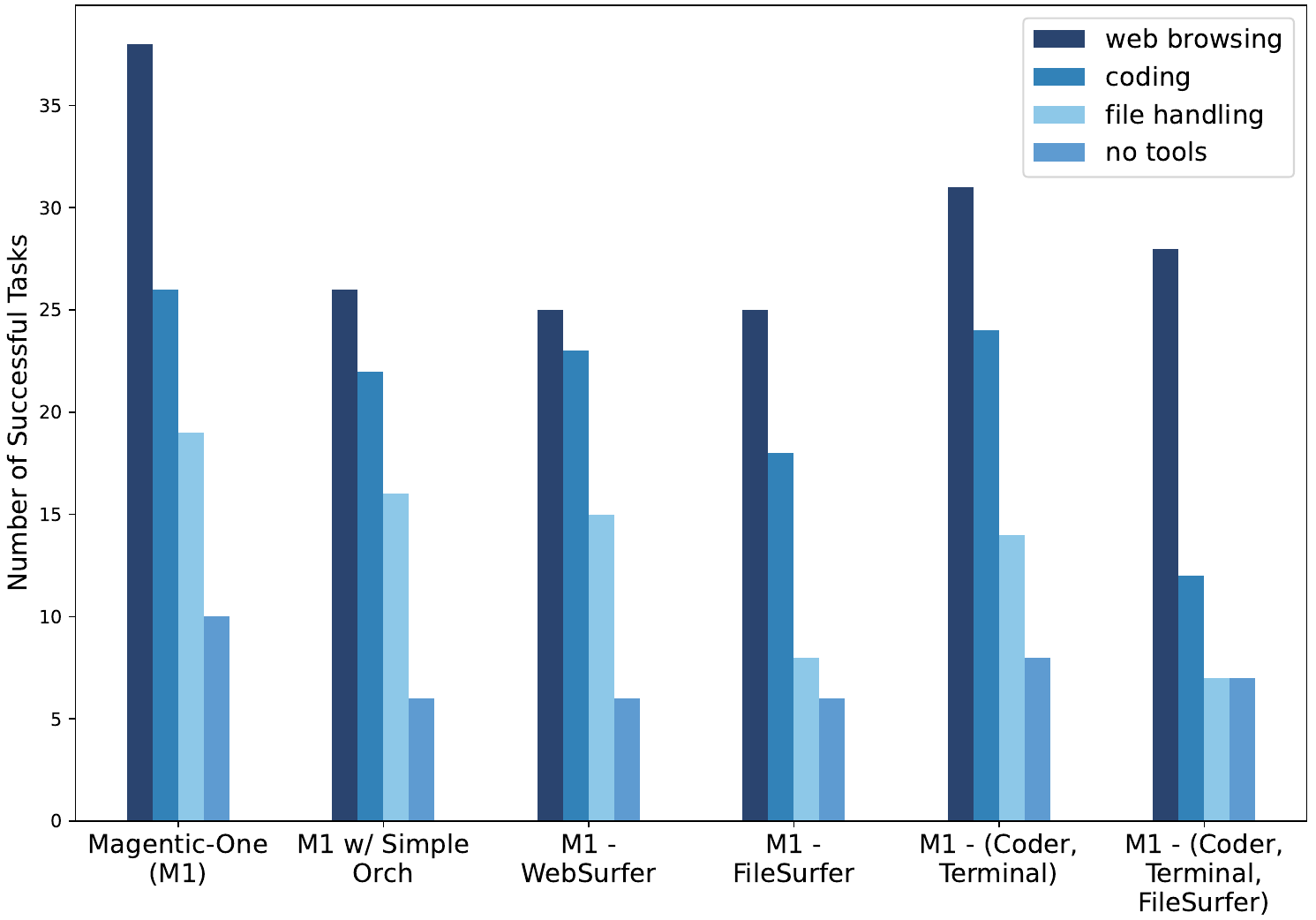}
    \caption{Performance by Capabilities}
    \label{fig:ablations_tool}
  \end{subfigure}
  \caption{Performance of different ablations of \team{} (GPT-4o) on the GAIA development set measuring the number of correct tasks. In the first ablation we replace the Orchestrator with a simple Orchestrator. In the following ablations we remove individual agents from \team{} denoted by ``-agent". The ablations show that all agents are essential to achieve the best performance.}
  \label{fig:ablations}
\end{figure}

\paragraph{Results.} In Figure \ref{fig:ablations_level}, we show the performance of different ablations of \team on the GAIA validation set broken down by difficulty level. We find that the Orchestrator's ledgers are important to \team's performance: without the full ledgers, performance drops by 31\%. 
Likewise, we find that all four worker agents are important: removing any single agent reduces performance by between 21\% (Coder, Executor) to 39\% (FileSurfer). For instance, the FileSurfer is essential for the largest GAIA category, evel 2, where many questions include file attachments. On the other hand, the WebSurfer is most essential for level 1 tasks. 

Figure \ref{fig:ablations_tool} shows ablation results broken down by required capabilities. In most cases, removing an agent from \team results in a decrease in team performance on tasks requiring corresponding capabilities. For example, \team with the FileSurfer removed shows the worst performance on tasks requiring file handling. Similarly, \team without the WebSurfer performs worst on tasks requiring web browsing.  

Interestingly, through qualitative analysis of the ablation logs, we found several cases where the \team agents compensated for missing capabilities in creative ways. For example, when the Coder and ComputerTerminal agents were not available for a task that was expected to require running code, the remaining agents solved the task by having the FileSurfer read and reason over the code to predict the answer. In another example, when the FileSurfer was unavailable for a task requiring reading contents of a pdf file, the remaining agents instead attempted to find an online pdf viewer to solve the task.  

\subsection{Error Analysis}

As a final element of evaluation, we conducted an analysis to better understand \team's current failure modes.

\paragraph{Approach.} As \team works to solve tasks, it produces extremely rich and detailed logs. Manual inspection of these logs often reveals mistakes, missed opportunities, dead-ends, and run-time errors encountered by the agents. Many of these issues are systematic, suggesting opportunities where the team could be improved. These opportunities could exist even when the agents successfully complete a task e.g., because of suboptimal behavior. However, manual inspection of these lengthy logs is slow and laborious, and scaling this manual labor to a large number of logs can become cost-prohibitive. 

To address this, we opted to automate log analysis using LLMs. The general problem here is to automate the process of {\em qualitative coding}, i.e., automatically discovering major themes in errors and inefficiencies observed in the logs. We implemented a multi-phase approach to accomplish this. For each task, we use GPT-4o to distill the team logs into a detailed postmortem document, which seeks to identify the root cause of failure, along with any contributing factors. These will serve as the basis for analysis. 

Each root-cause document is then automatically assigned a few descriptive codes (aka labels) using GPT-4o. With no pre-defined code book, there is initially a high diversity of codes across documents. After generating these initial codes, the next step is to group them into batches, with each batch being sent to  GPT-4o for clustering. This step merges similar codes into a more consolidated set. The process of consolidating and refining the codes is repeated iteratively, either until the codes stabilize or a maximum number of iterations is reached. 

We used 200 random samples of logs to bootstrap these codes and then once the final set of codes is determined, it is applied to the entire set of documents. 

\begin{figure}[!th]
    \centering
    \begin{subfigure}{0.8\linewidth}
        \centering
        \includegraphics[width=\linewidth]{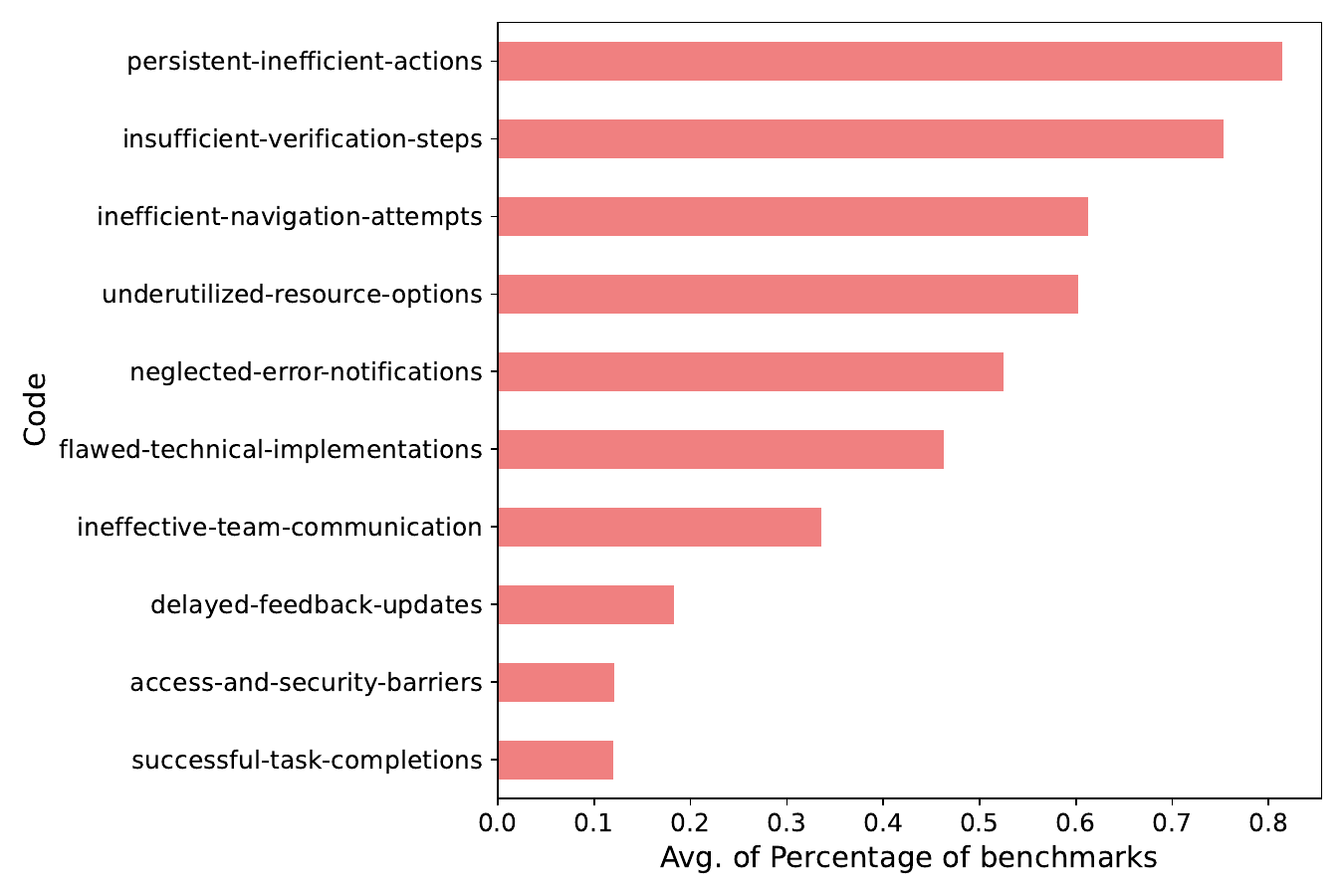}
        \caption{Distribution of error codes obtained by the automated analysis of \team's behavior as observed in the logs of the validation examples across all benchmarks studied.}
        \label{fig:errors}
    \end{subfigure}
    \\
    \begin{subfigure}{\linewidth}
        \centering
        \includegraphics[width=\linewidth]{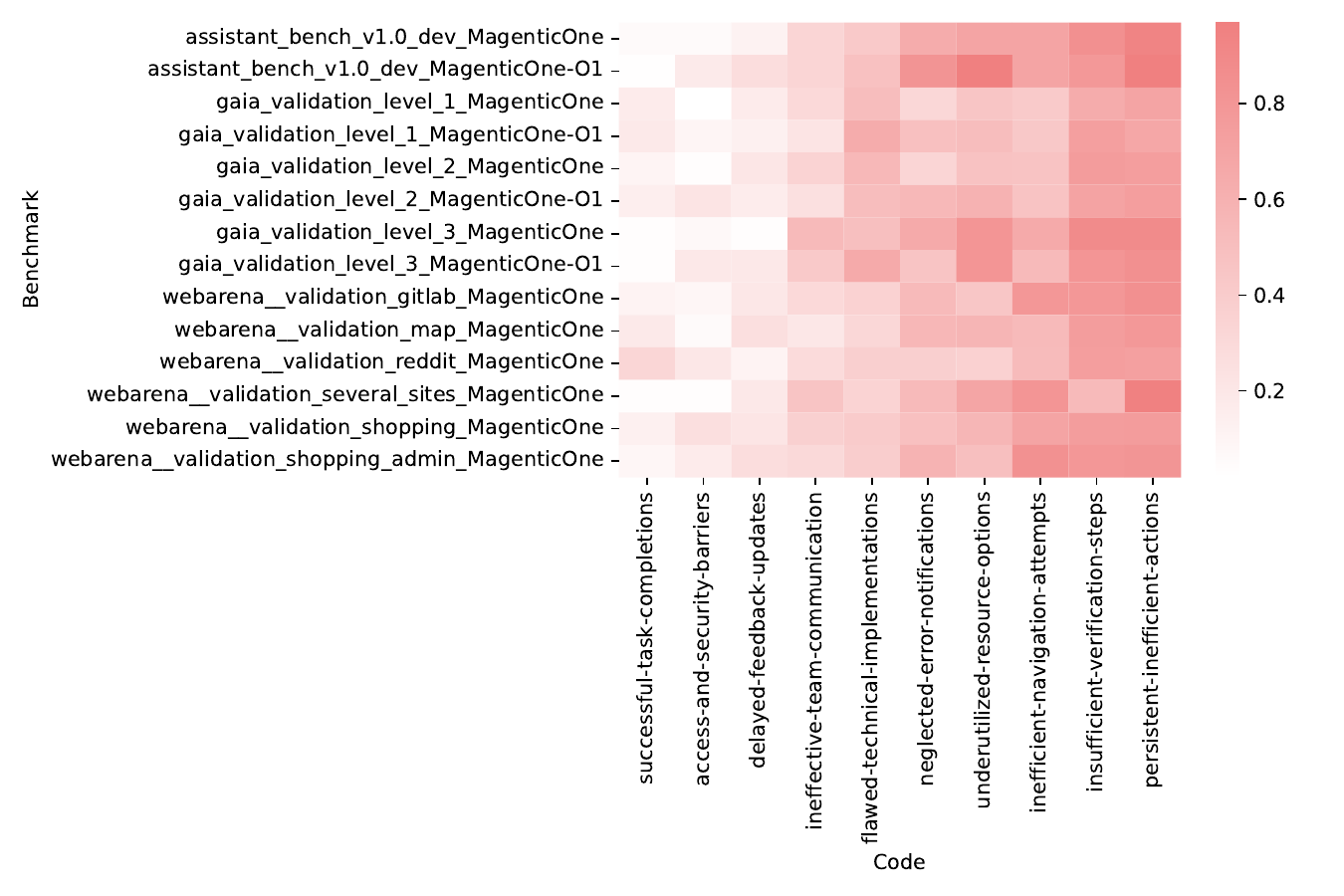}
        \caption{Heatmap of the error codes obtained by the automated analysis of \team's behavior as observed in the logs of the validation examples across all benchmarks studied.}
        \label{fig:confusion}
    \end{subfigure}
    \caption{Error analysis of \team's behavior.}
    \label{fig:error_analysis}
\end{figure}

\paragraph{Results.} Figure~\ref{fig:errors} shows the distribution of error codes that were automatically discovered by this approach for both versions of \team on the combined validation sets of all benchmarks. The codes are sorted by occurrence. Here we describe the top three codes. The details of all the codes, their definitions, and examples are available in Appendix~\ref{sec:appendix:error}. 

The most common code, persistent-inefficient-actions, refers to scenarios where agents repeatedly engage in unproductive behaviors without adapting their strategies, despite encountering failures. This persistence in ineffective actions leads to delays and suboptimal task outcomes. For instance, agents might continuously attempt the same unsuccessful web searches without modifying their queries or repeatedly access incorrect data sets without making necessary adjustments, resulting in wasted effort and time.

The second-most common code, insufficient-verification-steps, highlights situations where tasks are marked as complete without thorough validation of the data involved, leading to unreliable or erroneous results. Essential checks are bypassed, causing assumptions about data integrity that may not hold true. An example of this would be accepting final outputs without verifying their correctness, which can introduce errors into downstream analysis or decision-making processes due to unchecked inaccuracies.

The third-most common code, inefficient-navigation-attempts is related to errors arising from incorrect or inefficient navigation, which result in missed targets or prolonged task completion times. Agents often misinterpret interface layouts, leading to unnecessary cycling through tabs or menus. For example, an agent might repeatedly click through multiple tabs to locate the 'Settings' page, causing delays. Similarly, incorrect clicks on navigation bars can prevent access to the correct configuration settings. Confusion over user interface design can lead agents back to the main menu instead of the required subsection, further delaying task completion. Additionally, agents might persistently access incorrect page links, resulting in significant delays in retrieving important data. This code underscores the need for better navigation strategies and interface design to enhance task efficiency.

Figure~\ref{fig:confusion} shows a heat map of the codes broken down by specific benchmarks and version of \team. The heatmap again shows the presence of two most common codes -- persistent-inefficient-actions and insufficient-verification-steps -- across all benchmarks. The code underutilized-resource-options, which refers to scenarios where agents fail to utilize available data, tools, or resources effectively, is also prevalent in the logs. This code indicates that agents may not be taking full advantage of the resources at their disposal, leading to inefficient task execution and unnecessary manual actions. Another code, inefficient-navigation-attempts, is especially prevalent in the logs from the WebArena benchmark, where agents may struggle with interpreting interface layouts and taking inefficient paths to complete tasks.

\section{Discussion}
In this section, we discuss
open questions regarding the design of multi-agent systems for complex-tasks (Sec.~\ref{sec:The Multi-Agent Paradigm}),
current limitations (Sec.~\ref{sec:Limitations}),
and risks and risk mitigation for agents that autonomously operate computers (Sec.~\ref{sec:Risks and Mitigations}).

%%%%%%%%%%%%%%%%%%%%%%%%%%%%%%%%%%%%%%%%%%%%%%
\subsection{The Multi-Agent Paradigm}
\label{sec:The Multi-Agent Paradigm}

At the core of \team is its multi-agent design.
We believe that this design is a principal contributing factor to \team's performance.
Indeed, we observe that most other top-performing systems also follow a multi-agent design (Sec.~\ref{sec:Results}).

We argue that, beyond performance, the multi-agent setup offers numerous other advantages over the single-agent setup, in terms of ease-of-development, cost, and raw performance. 
%These benefits arise from the following properties:
For example, organizing skills into distinct agents can simplify development, much like with object-oriented programming. The separation of concerns across agents allows developers to focus model choices, prompting strategies, and other parameters to align to specific tasks (e.g., the web surfer agent benefits from multi-modality and structured output, but need not worry about writing code). Similarly, agent modularity can increase agent re-usability and ease of extensibility, particularly when teams are carefully designed to enable a plug-and-play approach. For example, \team's design facilitates adapting the team's functional scope by simply adding or removing agents, without requiring modifications to other agents's prompts or the overall flows and orchestration strategy. In contrast, monolithic single-agent systems often rely on constrained workflows that can be difficult to adapt or extend.

As a consequence of such modularity, each agent can be implemented in a fashion best suited for its purpose. In this paper, we leveraged this diversity to incorporate the o1-preview model into some roles (e.g., the Coder, and the outer loop  of the Orchestrator), while relying on a general purpose multi-modal model (GPT-4o) for web and file surfing. Looking ahead, we see this approach being used to reduce reliance on large models --  whereas some subtasks might require the largest language models available, others (e.g., grounding actions in WebSurfer, or summarizing large files in FileSurfer) might be amenable to much smaller -- and thus cheaper -- models. Different subtasks may also require different modalities, and some subtasks might be offloaded to traditional, non-AI, tools (e.g., code execution, for which a standard code execution environment is both sufficient and necessary). By embracing this diversity, multi-agent systems can become more performant at lower costs.

Understanding and quantifying the empirical advantages of multi- vs.\ single-agent setups constitutes a key question for future research.
Moreover, many variants of the multi-agent setup are possible. Here we opted for a single, centralized control flow pattern, where the Orchestrator agent plans for, and invokes, specialized worker agents.
Many other patterns are conceivable.
For instance, we might consider less centralized control flows, such as a peer-to-peer setting where each agent decides on its own which other agent should take control next.
At the other end of the spectrum, we might consider an even more rigid control flow where the orchestrator follows its own plan strictly (e.g., by encoding it as an executable program), rather than simply maintaining the plan it its prompt for chain-of-thought prompting.
Determining which control flow works best for which tasks is of considerable theoretical and practical importance.

In addition to the above-mentioned control-flow considerations, an alternate design dimension relates to the axes along which work is divided. \team's design diverges from other recent examples of multi-agent systems in that agents take on functional or tool-based responsibilities (web browser, computer terminal, etc.), rather than role-based responsibilities analogous to human teams (planner, researcher, data analyst, critic, etc.). In our experience, tool-centric agents can provide a cleaner separation of concerns compared to role-based agents, and a cleaner path to re-usability and compositionality -- if a web browser is a generic, multi-purpose tool, then a capable WebSurfer agent may hope to be generic and multi-purpose as well. Conversely, role-based patterns may require multiple agents to have redundant capabilities, while each agent fills only highly-specialized niche roles. For example, both a researcher and data analyst agent may need to operate a web browser or write code to complete their assigned tasks. Future work should empirically compare the performance of teams built with function- and role-based agents and examine the impact of each approach on the ease of development and debugging.

%%%%%%%%%%%%%%%%%%%%%%%%%%%%%%%%%%%%%%%%%%%%%%
\subsection{Limitations}
\label{sec:Limitations}
%%%%%%%%%%%%%%%%%%%%%%%%%%%%%%%%%%%%%%%%%%%%%%

Our work necessarily comes with certain limitations, some of which affect today's state of the field in general, and some of which are specific to our solution:

\begin{itemize}
    \item \textbf{Accuracy-focused evaluation:} Similar to other state-of-the-art systems, \team{} was evaluated on benchmarks that consider only the accuracy or correctness of final results. While considerably easier and more convenient to measure, such evaluations overlook  important considerations such as cost, latency, user preference and user value \cite{kapoor2024aiagentsmatter}. For example, even a partially correct trajectory may be valuable \cite{dibia-etal-2023-aligning}, whereas a perfectly accurate answer, delivered too late or at too high cost, may have no, or even negative, value. Designing evaluation protocols that incorporate these considerations, and that include subjective or open-ended tasks where correctness is less clear, remains an ongoing open-challenge in this the field.  
    \item \textbf{High cost and latency:} Although it was not part of the formal evaluation of \team, we would be remiss to skip mention of cost and latency in any discussion of limitations \cite{kapoor2024aiagentsmatter}. \team requires dozens of iterations and LLM calls to solve most problems. The latency and cost of those calls can be prohibitive, incurring perhaps several US dollars, and tens of minutes per task. We believe we can reduce these costs through targeted application of smaller local models, for example to support tool use in FileSurfer and WebSurfer, or set-of-mark action grounding in WebSurfer. Adding human oversight and humans-in-the-loop, may also save costs by reducing the number of iterations incurred when agents are stuck and problem-solving. This remains an active and ongoing area of future research.
    \item \textbf{Limited modalities:} \team cannot currently process or navigate all modalities. For example, WebSurfer cannot watch online videos -- though it often compensates by consulting transcripts or captions. Likewise, FileSurfer operates by converting all documents to Markdown, making it impossible to answer questions about a document's figures, visual presentation style, or layout. Audio files are similarly processed through a speech transcription model, so no agents can answer questions about music, or non-speech content. Benchmarks like GAIA exercise each of these skills. We would expect both benchmark and general task performance to improve with expanded support of multi-modal content. Future options include expanding \team's WebSurfer and FileSurfer agents' multi-modal capabilities or adding Audio and VideoSurfer agents specialized in handling audio and video processing tasks to the \team team. The latter approach is most inline with the value proposition of the multi-agent paradigm around easing development and reuse.      
    \item \textbf{Limited action space:} While agents in \team{} are afforded tools for the most common actions, tooling is not comprehensive. This simplifies the task of action grounding, but can lead to paths that are impossible to execute. For instance, the WebSurfer agent cannot hover over items on a webpage, or drag and resize elements. This can be limiting when interacting with maps, for example. Likewise, FileSurfer cannot handle all document types, and the Coder and Computer Terminal agents cannot execute code that requires API keys, or access to external databased or computational resources. We expect this limitation to close, over time, from two directions: first, we expect tool use to standardize across the industry, greatly enriching the set of tools available to agents. Second, similar to WebSurfer, agents will become better able to use operating systems and applications, affording them access to a broad range of tools developed for people.
    \item \textbf{Limited coding capabilities:} The \team{} Coder agent is particularly simple: it writes a new standalone Python program in respose to each coding request. In cases, where a prior coding attempt requires debugging, the Coder must correct the code by outputting an entirely new code listing. This is clearly not ideal. Two important limitations arise from this design choice: first, the Coder is ill-suited to operate over existing complex, or multi-file, code bases. Overcoming this limitation will be necessary to be competitive on benchmarks like SWE-bench \cite{jimenez2024swebenchlanguagemodelsresolve}. Second, the Coder sometimes fails because it expects functions that it previously defined to be available later in the workflow. Migrating to a Jupyter Notebook-like design, where later invocations simply add cells to a notebook, might mitigate this particular issue. This capability is presently supported by the AutoGen library upon which \team is built, and should be explored further.
    \item \textbf{Fixed team membership:} Additionally, \team's composition is fixed to a common set of five agents: Orchestrator, WebSurfer, FileSurfer, Coder, and ComputerTerminal. When agents are not needed, they simply serve as a distraction to the Orchestrator, and this may lower performance. Conversely, when extra expertise might be needed, it is simply unavailable. We can easily imagine an alternative approach where agents are added or removed dynamically, based on the task need. 
    \item \textbf{Limited learning:} Finally, although \team can adapt its strategy based on trial and error within a single task attempt, such insights are discarded between tasks. We observe the direct consequences of this design in WebArena, where many problems share a common set of core sub-tasks (e.g., finding a particular thread or user profile). When competing on this benchmark, \team's agents need to discover and rediscover solutions to these sub-tasks over and over. This is exhausting and frustrating to watch, is highly prone to error, and can incur significant additional costs. Overcoming this limitation through long-term memory is a key direction for future research.
\end{itemize}

%%%%%%%%%%%%%%%%%%%%%%%%%%%%%%%%%%%%%%%%%%%%%%
\subsection{Risks and Mitigations}
\label{sec:Risks and Mitigations}
%%%%%%%%%%%%%%%%%%%%%%%%%%%%%%%%%%%%%%%%%%%%%%

The agents described in this paper interact with a digital world designed for, and inhabited by, humans. This carries inherent and undeniable risks. In our work we mitigate such risks by running all tasks in containers, leveraging synthetic environments like WebArena, choosing models with strong alignment and pre- and post-generation filtering, and by closely monitoring logs during and after execution. Nevertheless, we observed the agents attempt steps that would otherwise be risky. For example, during development, a mis-configuration prevented agents from successfully logging in to a particular WebArena website. The agents attempted to log in to that website until the repeated attempts caused the account to be temporarily suspended. The agents then attempted to reset the account's password. In other cases, agents recognized that the WebArena Postmill website was not Reddit, then directed agents to the real website to commence work -- this was ultimately blocked by network-layer restrictions we had put in place. Likewise, we observed cases where agents quickly accepted cookie agreements and website terms and conditions without any human involvement (though captchas were correctly refused). More worryingly, in a handful of cases -- and until prompted otherwise -- the agents occasionally attempted to recruit other humans for help (e.g., by posting to social media, emailing textbook authors, or, in one case, drafting a freedom of information request to a government entity). In each of these cases, the agents failed because they did not have access to the requisite tools or accounts, and/or were stopped by human observers. To this end, it is imperative that agents operate under a strict principle of least privilege, and maximum oversight.

In addition to these observed and mitigated risks, we can anticipate new risks from such agentic systems on the horizon. As an example, as agents operate on the public internet, it is possible that they are subject to the same phishing, social engineering, and misinformation attacks that target human web surfers. The fact that the WebSurfer only occasionally recognized that Postmill was not Reddit, in WebArena, lends credence to the concern that agents can be fooled. If agents are equipped with a user's personal information  -- for example, to complete tasks on their behalf --  then this could potentially put that information at risk. Moreover, we can imagine that such attacks may be made more reliable and effective if attackers anticipate agentic use, and seed external material with specially crafted instructions or prompt-injections. To address these challenges, we can imagine several mitigations such as increasing human oversight, equipping agents with tools to validate external information (e.g., checking for typo-squatting in URLs, and ensuring TLS certificates, etc.), and including phishing rejection examples and other web-savvy skills in a model's post-training and instruction tuning. 

Another cross-cutting mitigation we anticipate becoming important is equipping agents with an understanding of which actions are easily reversible, which are reversible with some effort, and which cannot be undone. As an example, deleting files, sending emails, and filing forms, are unlikely to be easily reversed. This concept is explored in some detail in \cite{zhang2024interaction}, which provides a compelling framework for considering agent safety. When faced with a high-cost or irreversible action, systems should be designed to pause, and to seek human input.

Recent research has also investigated how interactions between multiple agents, such as iterative requests, long contexts, or cascading errors, may impact the effectiveness of the existing model alignment and guardrails upon which we rely. For example, the crescendo multi-turn attack \cite{russinovich2024great}, operates by prompting the model with a benign request, the slowly escalates the requests, building a pattern of agent compliance, until finally asking for a response that would otherwise be refused. In a multi-agent system, it is possible that a malicious agent, or an unfortunate accident, could result in a similar pattern of escalation. Fortunately, strong pre- and post-filtering, on the prompts and responses, remain a reasonable mitigation to such risks for the short-term. Moving forward, we strongly encourage model alignment work to focus on multi-turn scenarios. We also believe that red-team exercises are imperative to identify and mitigate such risks. 

Finally, there are potential long-term societal impacts of agentic systems, such as the potential to deskill or replace workers, leading to potential economic disruption. We believe it is therefore critical to work towards designing systems that facilitate effective collaboration between people and agents, such that humans and agents working together can achieve more than agents working alone. 

\section{Conclusions}

In this work we introduced \team, a generalist multi-agent system for ad-hoc, open-ended, file- and web-based tasks. \team uses a multi-agent architecture with a lead Orchestrator agent that directs four other agents. The Orchestrator agent is able to plan, track progress, and recover from errors with a ledger-based  orchestration. The remaining agents each specializes in the operation of generally-useful tools such as web browsers, file browsers, and computer console terminals. We show that \team is statistically competitive with other state-of-the-art (SOTA) systems on three challenging benchmarks, demonstrating both strong performance and generalization. Additionally, we have open-sourced the implementation of \team, which includes a reference framework for event-driven agents using the AutoGen framework. Finally, we discussed the limitations of \team, and the risks introduced by generalist AI agents, together with possible mitigation. To this end, \team represents a significant development in agentic systems capable of solving open-ended tasks.

\bibliographystyle{abbrv}
\bibliography{ref}

\begin{thebibliography}{10}

\bibitem{abuelsaad2024agenteautonomouswebnavigation}
T.~Abuelsaad, D.~Akkil, P.~Dey, A.~Jagmohan, A.~Vempaty, and R.~Kokku.
\newblock Agent-e: From autonomous web navigation to foundational design principles in agentic systems, 2024.

\bibitem{babyagi}
BabyAGI.
\newblock Github | babyagi.
\newblock \url{https://github.com/yoheinakajima/babyagi}, 2023.

\bibitem{bonatti2024windowsagentarenaevaluating}
R.~Bonatti, D.~Zhao, F.~Bonacci, D.~Dupont, S.~Abdali, Y.~Li, Y.~Lu, J.~Wagle, K.~Koishida, A.~Bucker, L.~Jang, and Z.~Hui.
\newblock Windows agent arena: Evaluating multi-modal os agents at scale, 2024.

\bibitem{cao2024spider2vfarmultimodalagents}
R.~Cao, F.~Lei, H.~Wu, J.~Chen, Y.~Fu, H.~Gao, X.~Xiong, H.~Zhang, Y.~Mao, W.~Hu, T.~Xie, H.~Xu, D.~Zhang, S.~Wang, R.~Sun, P.~Yin, C.~Xiong, A.~Ni, Q.~Liu, V.~Zhong, L.~Chen, K.~Yu, and T.~Yu.
\newblock Spider2-v: How far are multimodal agents from automating data science and engineering workflows?, 2024.

\bibitem{chen2024treesearchusefulllm}
Z.~Chen, M.~White, R.~Mooney, A.~Payani, Y.~Su, and H.~Sun.
\newblock When is tree search useful for llm planning? it depends on the discriminator, 2024.

\bibitem{cheng2024exploring}
Y.~Cheng, C.~Zhang, Z.~Zhang, X.~Meng, S.~Hong, W.~Li, Z.~Wang, Z.~Wang, F.~Yin, J.~Zhao, et~al.
\newblock Exploring large language model based intelligent agents: Definitions, methods, and prospects.
\newblock {\em arXiv preprint arXiv:2401.03428}, 2024.

\bibitem{d2024marg}
M.~D'Arcy, T.~Hope, L.~Birnbaum, and D.~Downey.
\newblock Marg: Multi-agent review generation for scientific papers.
\newblock {\em arXiv preprint arXiv:2401.04259}, 2024.

\bibitem{deng2023mind2webgeneralistagentweb}
X.~Deng, Y.~Gu, B.~Zheng, S.~Chen, S.~Stevens, B.~Wang, H.~Sun, and Y.~Su.
\newblock Mind2web: Towards a generalist agent for the web, 2023.

\bibitem{dibia-etal-2023-aligning}
V.~Dibia, A.~Fourney, G.~Bansal, F.~Poursabzi-Sangdeh, H.~Liu, and S.~Amershi.
\newblock Aligning offline metrics and human judgments of value for code generation models.
\newblock In A.~Rogers, J.~Boyd-Graber, and N.~Okazaki, editors, {\em Findings of the Association for Computational Linguistics: ACL 2023}, pages 8516--8528, Toronto, Canada, July 2023. Association for Computational Linguistics.

\bibitem{drouin2024workarenacapablewebagents}
A.~Drouin, M.~Gasse, M.~Caccia, I.~H. Laradji, M.~D. Verme, T.~Marty, L.~Boisvert, M.~Thakkar, Q.~Cappart, D.~Vazquez, N.~Chapados, and A.~Lacoste.
\newblock Workarena: How capable are web agents at solving common knowledge work tasks?, 2024.

\bibitem{du2023improving}
Y.~Du, S.~Li, A.~Torralba, J.~B. Tenenbaum, and I.~Mordatch.
\newblock Improving factuality and reasoning in language models through multiagent debate.
\newblock {\em arXiv preprint arXiv:2305.14325}, 2023.

\bibitem{Grosz1999SharedPlans}
B.~J. Grosz and S.~Kraus.
\newblock The evolution of sharedplans.
\newblock In {\em Proceedings of the International Conference on Multi-Agent Systems}, 1999.

\bibitem{guo2024large}
T.~Guo, X.~Chen, Y.~Wang, R.~Chang, S.~Pei, N.~V. Chawla, O.~Wiest, and X.~Zhang.
\newblock Large language model based multi-agents: A survey of progress and challenges.
\newblock {\em arXiv preprint arXiv:2402.01680}, 2024.

\bibitem{he2024webvoyagerbuildingendtoendweb}
H.~He, W.~Yao, K.~Ma, W.~Yu, Y.~Dai, H.~Zhang, Z.~Lan, and D.~Yu.
\newblock Webvoyager: Building an end-to-end web agent with large multimodal models, 2024.

\bibitem{hong2023metagpt}
S.~Hong, X.~Zheng, J.~Chen, Y.~Cheng, C.~Zhang, Z.~Wang, S.~K.~S. Yau, Z.~Lin, L.~Zhou, C.~Ran, et~al.
\newblock Metagpt: Meta programming for multi-agent collaborative framework.
\newblock {\em arXiv preprint arXiv:2308.00352}, 2023.

\bibitem{Jennings1998Applications}
N.~R. Jennings and M.~Wooldridge.
\newblock Applications of intelligent agents.
\newblock In {\em Proceedings of the International Conference on Autonomous Agents}, 1998.

\bibitem{jimenez2024swebenchlanguagemodelsresolve}
C.~E. Jimenez, J.~Yang, A.~Wettig, S.~Yao, K.~Pei, O.~Press, and K.~Narasimhan.
\newblock Swe-bench: Can language models resolve real-world github issues?, 2024.

\bibitem{kapoor2024aiagentsmatter}
S.~Kapoor, B.~Stroebl, Z.~S. Siegel, N.~Nadgir, and A.~Narayanan.
\newblock Ai agents that matter, 2024.

\bibitem{koh2024treesearchlanguagemodel}
J.~Y. Koh, S.~McAleer, D.~Fried, and R.~Salakhutdinov.
\newblock Tree search for language model agents, 2024.

\bibitem{li2024websuite}
E.~Li and J.~Waldo.
\newblock Websuite: Systematically evaluating why web agents fail.
\newblock {\em arXiv preprint arXiv:2406.01623}, 2024.

\bibitem{li2023camel}
G.~Li, H.~A. A.~K. Hammoud, H.~Itani, D.~Khizbullin, and B.~Ghanem.
\newblock Camel: Communicative agents for "mind" exploration of large scale language model society, 2023.

\bibitem{liang-arxiv2023}
T.~Liang, Z.~He, W.~Jiao, X.~Wang, Y.~Wang, R.~Wang, Y.~Yang, Z.~Tu, and S.~Shi.
\newblock Encouraging divergent thinking in large language models through multi-agent debate, 2023.

\bibitem{liu2024visualwebbenchfarmultimodalllms}
J.~Liu, Y.~Song, B.~Y. Lin, W.~Lam, G.~Neubig, Y.~Li, and X.~Yue.
\newblock Visualwebbench: How far have multimodal llms evolved in web page understanding and grounding?, 2024.

\bibitem{liu-arxiv2024}
N.~Liu, L.~Chen, X.~Tian, W.~Zou, K.~Chen, and M.~Cui.
\newblock From llm to conversational agent: A memory enhanced architecture with fine-tuning of large language models.
\newblock {\em arXiv e-prints}, pages arXiv--2401, 2024.

\bibitem{liu2024agent}
Y.~Liu, S.~K. Lo, Q.~Lu, L.~Zhu, D.~Zhao, X.~Xu, S.~Harrer, and J.~Whittle.
\newblock Agent design pattern catalogue: A collection of architectural patterns for foundation model based agents.
\newblock {\em arXiv preprint arXiv:2405.10467}, 2024.

\bibitem{lu2024ai}
C.~Lu, C.~Lu, R.~T. Lange, J.~Foerster, J.~Clune, and D.~Ha.
\newblock The ai scientist: Towards fully automated open-ended scientific discovery.
\newblock {\em arXiv preprint arXiv:2408.06292}, 2024.

\bibitem{masterman2024landscape}
T.~Masterman, S.~Besen, M.~Sawtell, and A.~Chao.
\newblock The landscape of emerging ai agent architectures for reasoning, planning, and tool calling: A survey.
\newblock {\em arXiv preprint arXiv:2404.11584}, 2024.

\bibitem{Messing2002AnIT}
B.~Messing.
\newblock An introduction to multiagent systems.
\newblock {\em K{\"u}nstliche Intell.}, 17:58--, 2002.

\bibitem{mialon2023gaiabenchmarkgeneralai}
G.~Mialon, C.~Fourrier, C.~Swift, T.~Wolf, Y.~LeCun, and T.~Scialom.
\newblock Gaia: a benchmark for general ai assistants, 2023.

\bibitem{mialon-arxiv2023}
G.~Mialon, C.~Fourrier, C.~Swift, T.~Wolf, Y.~LeCun, and T.~Scialom.
\newblock Gaia: benchmark for general ai assistants.
\newblock {\em arXiv preprint arXiv:2311.12983}, 2023.

\bibitem{nakano2021webgpt}
R.~Nakano, J.~Hilton, S.~Balaji, J.~Wu, L.~Ouyang, C.~Kim, C.~Hesse, S.~Jain, V.~Kosaraju, W.~Saunders, et~al.
\newblock Webgpt: Browser-assisted question-answering with human feedback.
\newblock {\em arXiv preprint arXiv:2112.09332}, 2021.

\bibitem{russellnorvig1996}
N.~J. Nilsson.
\newblock Stuart russell and peter norvig, artificial intelligence: A modern approach.
\newblock {\em Artificial Intelligence}, 82:369--380, 1996.

\bibitem{openai2023gpt4}
OpenAI.
\newblock Gpt-4 technical report, 2023.

\bibitem{pan2024autonomousevaluationrefinementdigital}
J.~Pan, Y.~Zhang, N.~Tomlin, Y.~Zhou, S.~Levine, and A.~Suhr.
\newblock Autonomous evaluation and refinement of digital agents, 2024.

\bibitem{pan2024webcanvasbenchmarkingwebagents}
Y.~Pan, D.~Kong, S.~Zhou, C.~Cui, Y.~Leng, B.~Jiang, H.~Liu, Y.~Shang, S.~Zhou, T.~Wu, and Z.~Wu.
\newblock Webcanvas: Benchmarking web agents in online environments, 2024.

\bibitem{paranjape2023art}
B.~Paranjape, S.~Lundberg, S.~Singh, H.~Hajishirzi, L.~Zettlemoyer, and M.~T. Ribeiro.
\newblock Art: Automatic multi-step reasoning and tool-use for large language models.
\newblock {\em arXiv preprint arXiv:2303.09014}, 2023.

\bibitem{park2023generativeagentsinteractivesimulacra}
J.~S. Park, J.~C. O'Brien, C.~J. Cai, M.~R. Morris, P.~Liang, and M.~S. Bernstein.
\newblock Generative agents: Interactive simulacra of human behavior, 2023.

\bibitem{paul-etal-2024-refiner}
D.~Paul, M.~Ismayilzada, M.~Peyrard, B.~Borges, A.~Bosselut, R.~West, and B.~Faltings.
\newblock {REFINER}: Reasoning feedback on intermediate representations.
\newblock In Y.~Graham and M.~Purver, editors, {\em Proceedings of the 18th Conference of the European Chapter of the Association for Computational Linguistics (Volume 1: Long Papers)}, pages 1100--1126, St. Julian{'}s, Malta, Mar. 2024. Association for Computational Linguistics.

\bibitem{putta2024agentqadvancedreasoning}
P.~Putta, E.~Mills, N.~Garg, S.~Motwani, C.~Finn, D.~Garg, and R.~Rafailov.
\newblock Agent q: Advanced reasoning and learning for autonomous ai agents, 2024.

\bibitem{qin2023tool}
Y.~Qin, S.~Hu, Y.~Lin, W.~Chen, N.~Ding, G.~Cui, Z.~Zeng, Y.~Huang, C.~Xiao, C.~Han, Y.~R. Fung, Y.~Su, H.~Wang, C.~Qian, R.~Tian, K.~Zhu, S.~Liang, X.~Shen, B.~Xu, Z.~Zhang, Y.~Ye, B.~Li, Z.~Tang, J.~Yi, Y.~Zhu, Z.~Dai, L.~Yan, X.~Cong, Y.~Lu, W.~Zhao, Y.~Huang, J.~Yan, X.~Han, X.~Sun, D.~Li, J.~Phang, C.~Yang, T.~Wu, H.~Ji, Z.~Liu, and M.~Sun.
\newblock Tool learning with foundation models, 2023.

\bibitem{qin2023toolllm}
Y.~Qin, S.~Liang, Y.~Ye, K.~Zhu, L.~Yan, Y.~Lu, Y.~Lin, X.~Cong, X.~Tang, B.~Qian, S.~Zhao, R.~Tian, R.~Xie, J.~Zhou, M.~Gerstein, D.~Li, Z.~Liu, and M.~Sun.
\newblock Toolllm: Facilitating large language models to master 16000+ real-world apis, 2023.

\bibitem{redcell_trase_2024}
{Red Cell Partners}.
\newblock Trase tops gaia leaderboard, 2024.

\bibitem{reed2022generalist}
S.~Reed, K.~Zolna, E.~Parisotto, S.~G. Colmenarejo, A.~Novikov, G.~Barth-Maron, M.~Gimenez, Y.~Sulsky, J.~Kay, J.~T. Springenberg, et~al.
\newblock A generalist agent.
\newblock {\em arXiv preprint arXiv:2205.06175}, 2022.

\bibitem{russinovich2024great}
M.~Russinovich, A.~Salem, and R.~Eldan.
\newblock Great, now write an article about that: The crescendo multi-turn llm jailbreak attack.
\newblock {\em arXiv preprint arXiv:2404.01833}, 2024.

\bibitem{Scerri2001AdjustableAI}
P.~Scerri, D.~V. Pynadath, and M.~Tambe.
\newblock Adjustable autonomy in real-world multi-agent environments.
\newblock In {\em International Conference on Autonomous Agents}, 2001.

\bibitem{schick-arxiv2023}
T.~Schick, J.~Dwivedi-Yu, R.~Dessì, R.~Raileanu, M.~Lomeli, L.~Zettlemoyer, N.~Cancedda, and T.~Scialom.
\newblock Toolformer: Language models can teach themselves to use tools, 2023.

\bibitem{shi2017world}
T.~Shi, A.~Karpathy, L.~Fan, J.~Hernandez, and P.~Liang.
\newblock World of bits: An open-domain platform for web-based agents.
\newblock In {\em International Conference on Machine Learning}. PMLR, 2017.

\bibitem{shinn2024reflexion}
N.~Shinn, F.~Cassano, A.~Gopinath, K.~Narasimhan, and S.~Yao.
\newblock Reflexion: Language agents with verbal reinforcement learning.
\newblock {\em Advances in Neural Information Processing Systems}, 36, 2024.

\bibitem{sodhi2024stepstackedllmpolicies}
P.~Sodhi, S.~R.~K. Branavan, Y.~Artzi, and R.~McDonald.
\newblock Step: Stacked llm policies for web actions, 2024.

\bibitem{song2024trialerrorexplorationbasedtrajectory}
Y.~Song, D.~Yin, X.~Yue, J.~Huang, S.~Li, and B.~Y. Lin.
\newblock Trial and error: Exploration-based trajectory optimization for llm agents, 2024.

\bibitem{MASaSurvey2000}
P.~Stone and M.~Veloso.
\newblock Multiagent systems: A survey from a machine learning perspective.
\newblock {\em Auton. Robots}, 8(3):345–383, June 2000.

\bibitem{talebirad2023multiagentcollaborationharnessingpower}
Y.~Talebirad and A.~Nadiri.
\newblock Multi-agent collaboration: Harnessing the power of intelligent llm agents, 2023.

\bibitem{Tambe1998AgentTeams}
M.~Tambe.
\newblock Implementing agent teams in dynamic multiagent environments.
\newblock {\em Appl. Artif. Intell.}, 12:189--210, 1998.

\bibitem{Wang2023survey}
L.~Wang, C.~Ma, X.~Feng, Z.~Zhang, H.~Yang, J.~Zhang, Z.~Chen, J.~Tang, X.~Chen, Y.~Lin, et~al.
\newblock A survey on large language model based autonomous agents.
\newblock {\em arXiv preprint arXiv:2308.11432}, 2023.

\bibitem{wang2024opendevinopenplatformai}
X.~Wang, B.~Li, Y.~Song, F.~F. Xu, X.~Tang, M.~Zhuge, J.~Pan, Y.~Song, B.~Li, J.~Singh, H.~H. Tran, F.~Li, R.~Ma, M.~Zheng, B.~Qian, Y.~Shao, N.~Muennighoff, Y.~Zhang, B.~Hui, J.~Lin, R.~Brennan, H.~Peng, H.~Ji, and G.~Neubig.
\newblock Opendevin: An open platform for ai software developers as generalist agents, 2024.

\bibitem{wang2024sibylsimpleeffectiveagent}
Y.~Wang, T.~Shen, L.~Liu, and J.~Xie.
\newblock Sibyl: Simple yet effective agent framework for complex real-world reasoning, 2024.

\bibitem{wang2024agentworkflowmemory}
Z.~Z. Wang, J.~Mao, D.~Fried, and G.~Neubig.
\newblock Agent workflow memory, 2024.

\bibitem{wei2022chain}
J.~Wei, X.~Wang, D.~Schuurmans, M.~Bosma, E.~Chi, Q.~Le, and D.~Zhou.
\newblock Chain of thought prompting elicits reasoning in large language models.
\newblock {\em arXiv preprint arXiv:2201.11903}, 2022.

\bibitem{Wooldridge1995}
M.~Wooldridge and N.~R. Jennings.
\newblock Intelligent agents: theory and practice.
\newblock {\em The Knowledge Engineering Review}, 10:115 -- 152, 1995.

\bibitem{wu2023autogen}
Q.~Wu, G.~Bansal, J.~Zhang, Y.~Wu, B.~Li, E.~Zhu, L.~Jiang, X.~Zhang, S.~Zhang, J.~Liu, A.~H. Awadallah, R.~W. White, D.~Burger, and C.~Wang.
\newblock Autogen: Enabling next-gen llm applications via multi-agent conversation framework.
\newblock In {\em COLM}, 2024.

\bibitem{wu2024oscopilotgeneralistcomputeragents}
Z.~Wu, C.~Han, Z.~Ding, Z.~Weng, Z.~Liu, S.~Yao, T.~Yu, and L.~Kong.
\newblock Os-copilot: Towards generalist computer agents with self-improvement, 2024.

\bibitem{xi2023risepotentiallargelanguage}
Z.~Xi, W.~Chen, X.~Guo, W.~He, Y.~Ding, B.~Hong, M.~Zhang, J.~Wang, S.~Jin, E.~Zhou, R.~Zheng, X.~Fan, X.~Wang, L.~Xiong, Y.~Zhou, W.~Wang, C.~Jiang, Y.~Zou, X.~Liu, Z.~Yin, S.~Dou, R.~Weng, W.~Cheng, Q.~Zhang, W.~Qin, Y.~Zheng, X.~Qiu, X.~Huang, and T.~Gui.
\newblock The rise and potential of large language model based agents: A survey, 2023.

\bibitem{xia2024agentless}
C.~S. Xia, Y.~Deng, S.~Dunn, and L.~Zhang.
\newblock Agentless: Demystifying llm-based software engineering agents.
\newblock {\em arXiv preprint arXiv:2407.01489}, 2024.

\bibitem{xie2024osworldbenchmarkingmultimodalagents}
T.~Xie, D.~Zhang, J.~Chen, X.~Li, S.~Zhao, R.~Cao, T.~J. Hua, Z.~Cheng, D.~Shin, F.~Lei, Y.~Liu, Y.~Xu, S.~Zhou, S.~Savarese, C.~Xiong, V.~Zhong, and T.~Yu.
\newblock Osworld: Benchmarking multimodal agents for open-ended tasks in real computer environments, 2024.

\bibitem{yang2023autogptonlinedecisionmaking}
H.~Yang, S.~Yue, and Y.~He.
\newblock Auto-gpt for online decision making: Benchmarks and additional opinions, 2023.

\bibitem{yang2024swe}
J.~Yang, C.~E. Jimenez, A.~Wettig, K.~Lieret, S.~Yao, K.~Narasimhan, and O.~Press.
\newblock Swe-agent: Agent-computer interfaces enable automated software engineering.
\newblock {\em arXiv preprint arXiv:2405.15793}, 2024.

\bibitem{yang2023set}
J.~Yang, H.~Zhang, F.~Li, X.~Zou, C.~Li, and J.~Gao.
\newblock Set-of-mark prompting unleashes extraordinary visual grounding in gpt-4v.
\newblock {\em arXiv preprint arXiv:2310.11441}, 2023.

\bibitem{yao2023webshopscalablerealworldweb}
S.~Yao, H.~Chen, J.~Yang, and K.~Narasimhan.
\newblock Webshop: Towards scalable real-world web interaction with grounded language agents, 2023.

\bibitem{yao2023treethoughtsdeliberateproblem}
S.~Yao, D.~Yu, J.~Zhao, I.~Shafran, T.~L. Griffiths, Y.~Cao, and K.~Narasimhan.
\newblock Tree of thoughts: Deliberate problem solving with large language models, 2023.

\bibitem{yao-iclr2023}
S.~Yao, J.~Zhao, D.~Yu, N.~Du, I.~Shafran, K.~Narasimhan, and Y.~Cao.
\newblock React: Synergizing reasoning and acting in language models.
\newblock In {\em International Conference on Learning Representations (ICLR)}, 2023.

\bibitem{yoran2024assistantbenchwebagentssolve}
O.~Yoran, S.~J. Amouyal, C.~Malaviya, B.~Bogin, O.~Press, and J.~Berant.
\newblock Assistantbench: Can web agents solve realistic and time-consuming tasks?, 2024.

\bibitem{zeng2023agenttuningenablinggeneralizedagent}
A.~Zeng, M.~Liu, R.~Lu, B.~Wang, X.~Liu, Y.~Dong, and J.~Tang.
\newblock Agenttuning: Enabling generalized agent abilities for llms, 2023.

\bibitem{zhang2024ufouifocusedagentwindows}
C.~Zhang, L.~Li, S.~He, X.~Zhang, B.~Qiao, S.~Qin, M.~Ma, Y.~Kang, Q.~Lin, S.~Rajmohan, D.~Zhang, and Q.~Zhang.
\newblock Ufo: A ui-focused agent for windows os interaction, 2024.

\bibitem{zhang2024cognitivekernelopensourceagent}
H.~Zhang, X.~Pan, H.~Wang, K.~Ma, W.~Yu, and D.~Yu.
\newblock Cognitive kernel: An open-source agent system towards generalist autopilots, 2024.

\bibitem{zhang2024webpilotversatileautonomousmultiagent}
Y.~Zhang, Z.~Ma, Y.~Ma, Z.~Han, Y.~Wu, and V.~Tresp.
\newblock Webpilot: A versatile and autonomous multi-agent system for web task execution with strategic exploration, 2024.

\bibitem{zhang2024autocoderover}
Y.~Zhang, H.~Ruan, Z.~Fan, and A.~Roychoudhury.
\newblock Autocoderover: Autonomous program improvement.
\newblock In {\em Proceedings of the 33rd ACM SIGSOFT International Symposium on Software Testing and Analysis}, pages 1592--1604, 2024.

\bibitem{zhang2024lookscreensmultimodalchainofaction}
Z.~Zhang and A.~Zhang.
\newblock You only look at screens: Multimodal chain-of-action agents, 2024.

\bibitem{zhang2024interaction}
Z.~J. Zhang, E.~Schoop, J.~Nichols, A.~Mahajan, and A.~Swearngin.
\newblock From interaction to impact: Towards safer ai agents through understanding and evaluating ui operation impacts.
\newblock {\em arXiv preprint arXiv:2410.09006}, 2024.

\bibitem{zhou2024webarenarealisticwebenvironment}
S.~Zhou, F.~F. Xu, H.~Zhu, X.~Zhou, R.~Lo, A.~Sridhar, X.~Cheng, T.~Ou, Y.~Bisk, D.~Fried, U.~Alon, and G.~Neubig.
\newblock Webarena: A realistic web environment for building autonomous agents, 2024.

\end{thebibliography}

\newpage 
\appendix
\section*{Appendix}

\section{Statistical Methodology}\label{apx:stats_results}

In Table \ref{tab:all_results}, we report the mean and an error bar for each reported method on the three benchmarks. To obtain the error bar we used a simple Wald 95\% confidence interval for the proportion. The Wald confidence interval assumes a normal approximation for the sample mean which is only valid for larger sample sizes and is only accurate for proportions not near 0 or 1. Our application meets these criteria: the smallest evaluated test set consists of 181 samples, and all reported results hover around 30\% with the exception of GPT-4. We also computed confidence intervals using the Wilson interval and found similar results (though Wilson intervals are not symmetric). For simplicity, we report only Wald intervals here. 

We also report the results of a statistical test comparing \team (GPT-4o, o1) to each reported method. We used the z-test to compare the accuracy of \team to each baseline in Tabe \ref{tab:all_results}. The z-test for proportions is the only feasible test in our setting because we only have the mean accuracy and not results on each example (the task-level test set results are hidden by the benchamrks). Therefore, we cannot apply McNemar's Test or a pairwise t-test. The limitation of the z-test is that it ignores pairing of the data and is generally conservative. We hope future benchmark leaderboards can release confidence intervals in addition to the reported mean.

\section{Capability to Category Mapping}\label{apx:capability_mapping}

Figure \ref{fig:ablations} (right) shows task performance results broken down by capabilities required. These capabilities are based on the capabilities human annotators reported as needed to solve tasks in the GAIA benchmark dataset \cite{mialon2023gaiabenchmarkgeneralai}. We organized and re-coded these annotations into the following categories to better reflect the roles of \team's agents:

\begin{itemize}
    \item \textbf{Web browsing:} capabilities related to searching and browsing the web. Examples: \texttt{web browser}, \texttt{search engine}, \texttt{maps}, \texttt{access to internet archives}
    \item \textbf{Coding:} capabilities related to coding and execution. Examples: \texttt{Coding}, \texttt{Python}, \texttt{calculator}, \texttt{audio/video processing}, \texttt{text processing}, \texttt{natural language processing}
    \item \textbf{File handling:} capabilities related to handling diverse file types. Examples: \texttt{Pdf viewer}, \texttt{Word}, \texttt{Excel}, \texttt{Powerpoint file access}, \texttt{CSV file access}, \texttt{XML file access}
    \item \textbf{No tools:} capabilities that can be performed inherently by agents, without tool-use, using multi-modal models. Examples: \texttt{Image recognition}, \texttt{OCR}, \texttt{Computer vision}, \texttt{color recognition}, \texttt{extracting text from images}
\end{itemize}

\newpage
\section{Error Analysis Code Book}
\label{sec:appendix:error}

In this section, we provide the final codes from the automated analysis of \team's behavior in the validation logs across all benchmarks. The codes are sorted by how often they appeared in the samples they were taken from. For each code, we include a definition and summaries of examples from the logs that were assigned that code.

{\small
\begin{longtable}{p{0.2\textwidth} p{0.3\textwidth} p{0.37\textwidth}}
\hline
\textbf{Name} & \textbf{Definition} & \textbf{Examples} \\
\hline
persistent-inefficient-actions &
Agents engaged in unproductive patterns without adapting despite facing failures. Ineffective strategies persisted, leading to delays and insufficient task outcomes. &
\begin{itemize}
  \item Agents clicked the same webpage sections multiple times, achieving no advancement in information retrieval.
  \item Agents unnecessarily engaged in general web searches instead of focusing on specified database tools.
  \item WebSurfer continued failing searches with no query modification, ignoring unsuccessful outcomes.
  \item The orchestrator commanded agents to use a flawed path repeatedly, indicating ongoing cycles without modification.
  \item Unaltered processes accessed incorrect data sets, consuming resources without advancing task goals.
\end{itemize}
\\ \hline

insufficient-verification-steps &
Tasks were marked complete without thorough data validation, leading to unreliable outcomes. Essential checks were skipped, resulting in erroneous assumptions about data integrity. &
\begin{itemize}
  \item Final outputs were accepted without validating data correctness, leading to potential errors.
  \item An orchestrator concluded a task although necessary criteria were unverified, risking incomplete achievement.
  \item A dataset's verification steps were skipped, causing unaddressed errors in downstream analysis.
  \item Document scans lacked quality checks before storing, leading to unreliable information in reports.
  \item Data interpretation lacked cross-verification assurances, uncovering gaps in computation assurances.
\end{itemize}
\\ \hline

underutilized-resource-options &
Agents consistently did not utilize available data, tools, or resources effectively. This resulted in inefficient task execution and repeated manual actions, even when automation was an option. &
\begin{itemize}
  \item Agents failed to integrate accessible descriptions, opting for redundant manual inputs despite metadata availability.
  \item Manual downloads persisted when FileSurfer was available for rapid document retrieval, unnecessarily extending the task.
  \item Available APIs were bypassed in favor of manual approaches, extending the task duration unnecessarily.
  \item Advanced search functions were overlooked, leading to reliance on broad manual explorations, slowing down the process.
  \item Complex data visualization was attempted with basic charting tools instead of comprehensive graphing tools.
\end{itemize}
\\ \hline

inefficient-navigation-attempts &
Errors occurred because of incorrect or inefficient navigation, leading to missed targets or prolonged task completion. Agents misinterpreted interface layouts and took inefficient paths. &
\begin{itemize}
  \item The agent mistakenly cycled through multiple tabs to find the 'Settings' page, resulting in delayed task progress.
  \item Incorrectly clicked navigation bars led to a user failing to access the proper configuration settings.
  \item Confusion over the UI design led an agent back to the main menu instead of the subsection needed for task completion.
  \item Cycling through history tabs resulted in missed current transaction logs.
  \item The agent persistently accessed the wrong page links, causing delays in retrieving important data.
\end{itemize}
\\ \hline

ineffective-team-communication &
Agents did not communicate information or direction effectively, causing task overlap and confusion. Miscommunications led to redundant work and uncoordinated actions. &
\begin{itemize}
  \item Two agents simultaneously accessed order histories due to unclear task division, resulting in duplicated efforts.
  \item Agents struggled to understand task requirements due to incomplete directive descriptions.
  \item The WebSurfer and FileSurfer failed to communicate their findings, leaving the Orchestrator to make decisions based on incomplete information.
  \item Repeated task restarts occurred because instructions were too vague, leading to incorrect interpretations.
  \item Agents observed unclear role demarcations that caused them to erroneously overwrite each other's progress.
\end{itemize}
\\ \hline

neglected-error-notifications &
Agents ignored known errors or warnings, allowing issues to persist and recur. This resulted in repeated inefficiencies and bottlenecks in task execution. &
\begin{itemize}
  \item Repeated 'ValueError' messages were ignored, allowing the underlying issue to continue unaddressed.
  \item Agents accessed 'hotel booking' pages multiple times despite 'No results found' errors repeatedly visible.
  \item System errors were logged without corrective attempts, leading to prolonged delays in task completion.
  \item Unresolved input errors led to repeated submissions of the same query form without variations.
  \item Despite validation warnings, agents continued with deprecated approaches, leading to unresolved problems.
\end{itemize}
\\ \hline

flawed-technical-implementations &
Tasks faced issues due to incorrect application of technical logic or processes. Misapplied techniques led to errors and inefficiencies. &
\begin{itemize}
  \item Agents encountered syntax errors in scripts due to incorrect indentation, halting task progress.
  \item Repeated runtime errors occurred as agents submitted misformatted queries without cross-verifying syntax.
  \item In accessing web modules, agents misapplied parameter mappings, causing non-responsive functions.
  \item Misinterpretation of calculation steps led to varying billing inconsistencies within an ecommerce task.
  \item Agents faced challenges in aligning sales data, resulting in flawed revenue projections.
\end{itemize}
\\ \hline

access-and-security-barriers &
Tasks were hindered due to security or access restrictions, affecting data integrity. Agents struggled with authentication, allowing unauthorized access or limiting task completion. &
\begin{itemize}
  \item Visible password fields were used, risking data exposure to unauthorized parties.
  \item Repeated attempts to submit forms were met with unresolved errors blocking access to data queries.
  \item Token visibility in the logs suggested possible unauthorized data access due to weak credential management.
  \item Insufficient password policies allowed repeated login attempts without user notifications, risking security.
  \item Repeated security filter alerts appeared as agents attempted to access restricted areas.
\end{itemize}
\\ \hline

delayed-feedback-updates &
There was a delay in communicating task progress or results, causing confusion and hindering coordination. Timeliness in updates was lacking, leaving tasks in ambiguity. &
\begin{itemize}
  \item After WebSurfer confirmed a task, the orchestrator delayed updating the user, leading to confusion.
  \item Notification of successful strategy analysis reached the user well after execution, causing temporary uncertainty.
  \item Despite quick report generation, notifications lagged behind, leading users to question task progression.
  \item Task completions were assumed due to lagged updates in user notifications post-webpage completions.
  \item Feedback on document completions took significantly longer to communicate to concerned tasks than average.
\end{itemize}
\\ \hline

successful-task-completions &
Tasks were completed smoothly with no errors, meeting all objectives efficiently. Agents achieved success through coordinated efforts, correct usage of resources, and thorough verification. &
\begin{itemize}
  \item During the data migration task, all user details were accurately uploaded and verified without issues.
  \item Correct transaction processing led to flawless financial reconciliation, with all figures matching expectations.
  \item Backend updates improved processing speeds, which performance tests later verified as positive and expected.
  \item Users experienced smooth account creations, following flawless processes without discrepancies or errors.
  \item Revisions in manuscript drafts met all editorial instructions perfectly, with edits applied and aligned as needed.
\end{itemize}
\\ \hline

\end{longtable}

}

\end{document}